\documentclass[pdflatex,sn-mathphys,Numbered]{sn-jnl}
\usepackage{array}

\usepackage{lmodern}      % scalable fonts (silences CM 'size not available' warnings)
\usepackage{graphicx}
\usepackage{amsmath,amssymb,amsfonts}
\usepackage{booktabs}
\usepackage{multirow}
\usepackage{makecell}
\usepackage{xcolor}
\usepackage{hyperref}
\usepackage{url}
\usepackage{algorithm}
\usepackage{algorithmicx}
\usepackage{algpseudocode}
\usepackage{tikz}
\usepackage{pgfplots}
\usepackage{pgfplotstable}
\pgfplotsset{compat=1.18}
\usepgfplotslibrary{groupplots}
\usepackage{rotating}   % for sidewaystable if needed

\definecolor{steelblue}{RGB}{70,130,180}
\definecolor{crimson}{RGB}{220,20,60}

\usepackage{placeins}                 % \FloatBarrier
\begin{document}

\title[PANDA: Prototype-Anchored Alignment for Partially Unpaired Multimodal Learning]{PANDA: Prototype-Anchored Alignment for Partially Unpaired
Multimodal Learning, with Applications to Alzheimer’s MRI and TCGA Pathology}

\author*[1]{\fnm{Sheethal} \sur{Bhat}}\email{sheethal.bhat@fau.de}
\author[1]{\fnm{Mahfuzur Rahman} \sur{Chowdhury}}
\author[1]{\fnm{Paula Andrea} \sur{P\'erez-Toro}}
\author[2,4]{\fnm{Stephan} \sur{Wunderlich}}
\author[3]{\fnm{Rose Dawn} \sur{Bharat}}
\author[1]{\fnm{Siming} \sur{Bayer}}
\author[1]{\fnm{Andreas} \sur{Maier}}

\affil*[1]{\orgdiv{Pattern Recognition Lab},
\orgname{Friedrich-Alexander-Universit\"at},
\orgaddress{\city{Erlangen-N\"urnberg}, \country{Germany}}}

\affil[2]{\orgname{Department of Neurology, Klinikum N\"urnberg, Paracelsus Medical University},
\orgaddress{\city{N\"urnberg}, \country{Germany}}}

\affil[3]{\orgname{National Institute of Mental Health and Neurosciences
(NIMHANS)}, \orgaddress{\city{Bengaluru}, \country{India}}}

\affil[4]{\orgname{Department of Radiology, LMU University Hospital, LMU Medizin, Ludwig-Maximilians-Universität München},
\orgaddress{\city{Munich}, \country{Germany}}}

\abstract{
Multimodal medical prediction frequently contends with incomplete pairing: auxiliary modalities with complementary signal are available for only a subset of subjects (or none) and cannot be assumed at deployment. We introduce PANDA (Prototype-Anchored Data Alignment), a two-stage framework that transfers auxiliary information to a primary-modality model without auxiliary inputs at inference. Two-stage training first derives class-specific auxiliary prototypes from the shared embedding, then aligns the primary encoder to these frozen prototypes. Because supervision is defined at the class-prototype level, PANDA accommodates arbitrary pairing rates, including zero subject overlap.

We evaluate PANDA on two applications. On a 1{,}021-subject multi-scanner ADNI cohort, we perform AD/CN classification with three auxiliary modalities at distinct pairing rates: tabular scores (44.8\%), FDG-PET (18.7\%), and external handwriting kinematics (0\% overlap). Relative to the same-backbone MRI-only baseline, PANDA improves AUC on both encoders, with a larger gain on the weaker one: on a transfer-pretrained MedicalNet backbone it attains AUC $0.868\pm0.020$ ($+7.9$\,pp over the MRI-only baseline) and reduces 1.5\,T CN false positives by 25.2\,pp, while on a stronger, fully trainable Conv5-FC3 backbone it reaches a SOTA AUC of $0.893$. On TCGA-Lung survival prediction from whole-slide images, using RNA-seq as auxiliary data, PANDA improves over WSI-only on 2-year OS (AUC $+3.5$\,pp) and Cox PH (C-index $+9.0$\,pts). It also outperforms full-fusion training, which underperforms WSI-only. PANDA requires no RNA at inference; wide confidence intervals on this smaller cohort keep the gains below conventional significance. Overall, PANDA provides a deployment-oriented mechanism for leveraging incomplete auxiliary modalities to improve primary-modality prediction.
}

\keywords{PANDA, Alzheimer's disease, MRI classification, unpaired multimodal learning,
prototype alignment, cross-domain generalisation.}

\maketitle

% ── Sections ─────────────────────────────────────────────────────────────────

% §1 Introduction
\section{Introduction}
\label{sec:intro}

Multimodal learning has become a standard approach in the general medical domain for prediction tasks~\cite{baltrusaitis2018multimodal,zhang2011multimodal,suk2014hierarchical}; however, most existing methods either assume that all modalities are jointly observed for every subject or rely on imputation and generative modelling to synthesize unobserved modalities.
In practical settings, this assumption is rarely met.
Auxiliary measurements that carry complementary signal, such as additional imaging, molecular assays, clinical scores, or behavioural markers, are typically available for only a subset of subjects in a large cohort.
Moreover, their availability varies from one modality to the
next, and they are frequently absent altogether at deployment. Discarding unpaired subjects or imputing missing modalities, the latter either
by generative synthesis~\cite{sharma2020missing,havaei2016hemis,hu2020knowledge}
or by reconstructing missing-modality features through Bayesian
meta-learning~\cite{ma2021smil}, are the usual responses, but both waste the large pool of
primary-modality data that \emph{is} available, either shrinking the training
set or injecting errors during imputation.

In this study, we investigate how a classifier constrained to a single primary modality at inference time can nonetheless leverage auxiliary modalities that are only partially paired with the primary modality during training, if paired at all.
To this end, we shift supervision from per-subject correspondences to class-level geometry~\cite{snell2017prototypical,radford2021learning}: auxiliary modalities are used solely to define fixed class prototypes, and the primary-modality encoder is aligned to these prototypes using all subjects, whether paired or unpaired.
We subsequently interpret this alignment in information-theoretic terms (Section~\ref{sec:infotheory}): it maximizes a lower bound on the mutual information between the primary encoder and the auxiliary modalities, thereby providing insight into why reduced pairing density need not degrade performance.

The approach is developed for medical imaging and evaluated in depth on structural MRI Alzheimer's disease (AD) classification, the primary application, enabling analyses of scanner robustness and disease-severity structure.
Structural MRI is a natural primary modality for the AD study because it is non-invasive, broadly available, and sensitive to the hippocampal and cortical atrophy that characterises disease progression~\cite{ADNIdataset,chen2019med3d,wen2020convolutional}. In the MRI/AD setting, real-world deployment additionally entails two interrelated challenges that are rarely addressed jointly.
Transfer is further demonstrated on a structurally unrelated second application: whole-slide-image cancer survival prediction with a genomic auxiliary modality~\cite{ilse2018attention,chen2021mcat}.

A first challenge is \emph{scanner heterogeneity}.
Clinical neuroimaging datasets are acquired across sites using scanners that differ in field strength (1.5\,T vs. 3\,T), manufacturer, and acquisition protocol.
Such variability induces systematic image-level domain shifts that can degrade classification performance~\cite{johnson2007adjusting,fortin2017harmonization,glocker2019machine}.
Scanner heterogeneity is frequently circumvented rather than addressed: evaluations are often restricted to a single field strength or protocol-matched cohorts, and head-to-head cross-scanner validation is uncommon, leaving scanner robustness insufficiently characterized~\cite{glocker2019machine,leming2022confounder,song2022reliability}.
In the ADNI~\cite{ADNIdataset} cohort, an MRI-only baseline performs substantially worse at 1.5\,T than at 3\,T, with elevated false-positive rates among 1.5\,T control subjects.
Although domain adaptation and harmonization methods can partially mitigate these shifts, they typically require scanner labels during training or paired acquisitions that are often unavailable in retrospective multi-site studies.

A second challenge is \emph{incomplete multimodal coverage}.
Alongside MRI, clinical cohorts routinely collect auxiliary measurements, including neuropsychological scores (MMSE, CDR, FAQ), FDG-PET, and digital biomarkers, each of which provides complementary disease-relevant information.
However, complete modality coverage is often unavailable at the subject level, and the paired subsets vary across modalities.
In ADNI~\cite{ADNIdataset}, cognitive scores and FDG-PET are each available for only a fraction of subjects. Moreover, some biomarker sources (e.g., digital handwriting kinematic assessments) are collected in entirely separate cohorts with no subject overlap.
Standard multimodal fusion methods either require complete pairing at both training and inference~\cite{zhang2011multimodal,suk2014hierarchical} or impute missing modalities~\cite{sharma2020missing,havaei2016hemis}; both are impractical at deployment and can introduce imputation-induced inductive bias.

These considerations motivate the hypothesis that, despite heterogeneity in raw feature space, the modalities encode a shared disease-discriminative structure, such that AD-versus-CN class separation is relatively modality-invariant.
Consequently, the class geometry induced by an auxiliary modality can serve as a supervisory target for the primary-modality encoder.
Aligning the primary encoder to this shared class-level geometry, rather than to per-subject correspondences, allows auxiliary modalities to shape the MRI representation for all subjects, whether paired or unpaired.

Specifically, this work introduces PANDA (Prototype-Anchored Data Alignment)\footnote{A preliminary version of this work was presented in abstract form at BAIOSPHERE MEDICAL~2026~\cite{bhat2026prototype}.}, a two-stage framework illustrated in Fig.~\ref{fig:overview}. 
In Stage~1, the MRI encoder and the auxiliary encoders are trained jointly on the available paired data to form a shared embedding, from which the auxiliary class prototypes are computed and then fixed (Fig.~\ref{fig:overview}, top). 
In Stage~2, the MRI encoder is trained on \emph{all} subjects to align with these fixed prototypes (middle). At inference time, the auxiliary encoders are discarded and each scan is classified from MRI alone (bottom). This formulation accommodates three qualitatively different pairing regimes within a single training procedure: subject-level tabular pairing (44.8\,\%), partial PET pairing (18.7\,\%), and an external handwriting-kinematics prototype with no ADNI~\cite{ADNIdataset} overlap (0\,\%).

On the ADNI~\cite{ADNIdataset} AD/CN cohort, the full model substantially outperforms an MRI-only baseline.
When the same prototype alignment is applied on a stronger, fully-trainable encoder, it attains the highest overall performance on this cohort (AUC 0.893), surpassing the strongest unimodal baseline (see Section~\ref{sec:backbone}).
A scanner-stratified analysis indicates that tabular alignment primarily reduces false-positive rates at 1.5\,T, whereas PET alignment improves discrimination at 3\,T.
A pairing-rate ablation yields a notable result: the joint Tab+PET anchor does not degrade as pairing is reduced, indicating that full pairing is unnecessary. In addition, a zero-shot analysis on held-out MCI subjects shows that an MRI-only baseline does not reliably preserve the ordinal severity continuum, whereas the joint prototype-aligned model does. A simple linear severity head, trained only on graded CN/AD scores and never on MCI, further exposes an explicit monotone CN$\to$AD severity axis without reducing AUC.
Finally, cross-domain validation on TCGA-Lung (whole-slide images with RNA-seq) indicates that the mechanism is not neuroimaging-specific, with directionally consistent gains that the smaller cohort leaves below significance. Thus, the main contributions of this paper are:
\begin{enumerate}
  \item \textbf{A framework for partially unpaired multimodal learning in medical imaging.}
    PANDA is a two-stage, prototype-anchored alignment framework that enables a primary-modality classifier to leverage auxiliary modalities across pairing
    regimes (subject-level, partial, or external-cohort with zero overlap) within a single training procedure, while requiring only the primary
    modality at inference. We demonstrate the framework on two medical-imaging domains---brain MRI (AD/CN classification) and lung-pathology whole-slide imaging (survival)---and scope the claim to these settings rather than asserting generality beyond the domains evaluated here.
  \item \textbf{Full pairing is not required.}
    A pairing-rate ablation showing that sparse heterogeneous pairing matches full pairing (within seed noise) for the joint anchor, with a mechanistic
    explanation based on class-prototype separation, a finding with direct implications for how multimodal cohorts are best collected.
  \item \textbf{Alzheimer's classification (primary application).} A systematic scanner-stratified evaluation on ADNI~\cite{ADNIdataset} showing that different auxiliary modalities target different sources of scanner bias (clinical-score prototypes reduce 1.5\,T false positives while PET prototypes improve 3\,T
    discrimination), and a zero-shot analysis on held-out MCI subjects showing that the joint prototype-aligned model preserves a monotone CN$\to$AD
    severity ordering that a single-modality baseline does not. A severity-head extension makes this axis explicit using graded CN/AD scores only; MCI
    never contributes a training gradient and binary AUC is unchanged.
  \item \textbf{Cancer survival (second, cross-domain application).}
    Validation on TCGA-Lung showing that the same prototype-transfer mechanism carries from 3D neuroimaging to gigapixel whole-slide imaging, exceeding
    full-fusion joint training despite using no RNA at inference.
\end{enumerate}

\section{Related Work}
\label{sec:related}

\subsection{MRI-Based AD Classification and Scanner Robustness}

Deep convolutional networks operating on volumetric T1-weighted MRI have demonstrated high AD/CN discrimination performance, although such results are typically reported under constrained evaluation conditions. Moreover, a substantial share of reported gains reflects methodological artefacts rather than genuine discrimination such as, subject-level data leakage between training and test partitions~\cite{wen2020convolutional,chowdhury2026robust}. 
Despite these caveats, transfer learning from large-scale medical imaging pretraining corpora (e.g., MedicalNet~\cite{chen2019med3d}) remains a dominant state-of-the-art (SOTA) approach in many AD/CN MRI benchmarks, as it lowers the sample complexity of AD-specific fine-tuning.
Regardless of architecture, a consistent limitation remains: sensitivity to scanner domain. Models trained on 3\,T ADNI scans degrade on 1.5\,T
acquisitions, where lower SNR and contrast differences systematically shift feature distributions~\cite{glocker2019machine}.

Two classes of approaches explicitly target this distribution shift, each with distinct assumptions and supervision requirements.
Batch-effect correction methods such as ComBat~\cite{johnson2007adjusting,fortin2017harmonization} and deep variants model scanner variation as a nuisance covariate and remove it via regression; they can be effective when scanner labels are available, but implicitly assume that biological effects are separable from scanner-induced variation, an assumption that can fail when the two are entangled.
Domain-adaptation methods instead reduce source--target discrepancy via adversarial training~\cite{ganin2016dann} or distribution matching~\cite{lee2019sliced}, but require access to target-domain scans during training and therefore do not naturally generalise to scanners unseen at development time.
Both families achieve robustness by introducing an explicit scanner model and relying on privileged information (scanner labels or target-domain data).

PANDA adopts an alternative strategy: instead of explicitly modelling scanner effects, it aligns the MRI representation to auxiliary modalities that are invariant to scanner acquisition by construction (e.g., cognitive scores, PET metabolism, or handwriting kinematics). This induces suppression of scanner-correlated variability as an implicit consequence of the alignment objective, without requiring scanner labels or access to target-domain scans during training or inference. Empirically, tabular-prototype alignment substantially reduces the false-positive rate on 1.5\,T scans.

\subsection{Multimodal Fusion and Missing Modality}

Multimodal fusion for AD has been studied extensively, ranging from early (feature-concatenation) to late (decision-level) fusion of MRI, PET, and
clinical scores~\cite{zhang2011multimodal,suk2014hierarchical}. Recent methods align multiple modalities for multiclass diagnosis, for example by combining
MRI, tau-PET, diffusion MRI, and cognitive scores via contrastive alignment and a tabular foundation model~\cite{huang2025multistage}.
Such methods typically assume that all modalities are available at inference, an assumption that is often violated in clinical cohorts.
Missing-modality methods relax this requirement via modality-dropout training~\cite{neverova2016moddrop}, generative imputation of absent modalities~\cite{sharma2020missing,havaei2016hemis}, knowledge distillation from a complete-modality teacher~\cite{hu2020knowledge,wang2020multimodalkd}, or Bayesian meta-learning of absent-modality features~\cite{ma2021smil}. In particular, Wang et al.~\cite{wang2020multimodalkd} address incomplete modalities via knowledge distillation from a complete-modality teacher; our distillation baseline follows this paradigm, whereas PANDA's prototype anchoring offers an alternative that does not require a fully paired teacher.
However, these approaches still require the auxiliary modality for at least a subset of subjects (or access to a complete-modality teacher), and therefore do not apply when the auxiliary cohort has \emph{zero} subject overlap, as in our handwriting regime ($r_\mathrm{HW}=0$). Prototype anchoring, by contrast, depends only on class-level geometry and can accommodate disjoint auxiliary modalities. In the federated setting, Le et al.~\cite{le2025crossmodal} propose cross-modal prototype-based alignment to handle severely missing modalities across clients; we adapt a similar prototype-anchored principle to the centralised, partially unpaired setting, in which auxiliary modalities can be sparsely paired or drawn from an entirely external cohort rather than distributed across federated clients.

Closer to our setting, recent AD-specific methods handle incomplete data without explicit imputation. Ou et al.~\cite{ou2024graph} learn a graph-embedded latent
space over all subjects, including those missing PET, and Liu et al.~\cite{liu2025progressive} align features across acquisition stages so that costly later-stage modalities need not be acquired at test time.
PANDA shares this deployment objective but aligns the primary encoder to \emph{frozen class-level prototypes} rather than to a fused latent space or per-subject stage features, enabling the use of an entirely external auxiliary cohort.
A separate line of work addresses \emph{partial pairing} by restricting multimodal training to the fully paired subset and discarding subjects missing any modality~\cite{suk2014hierarchical}. We show that this strategy is detrimental: a joint model trained only on the paired subset underperforms the unimodal baseline on both ADNI and TCGA (Tables~\ref{tab:main_results},~\ref{tab:tcga}), because the paired subset is a smaller and biased sample of the training distribution. PANDA avoids this failure mode by separating alignment (Stage~1, paired subjects) from classification (Stage~2, all subjects).

\subsection{Contrastive and Prototype Learning}

Self-supervised contrastive learning is widely used for representation learning, ranging from instance-discrimination frameworks~\cite{chen2020simclr} to supervised contrastive objectives that encourage within-class compactness in the embedding space~\cite{khosla2020supcon}.
Cross-modal contrastive learning based on the InfoNCE objective, exemplified by CLIP~\cite{radford2021learning} for image--text pairs, has also been applied in medical imaging for modality alignment~\cite{zhang2022convirt} and cross-modal retrieval.
In parallel, prototypical networks~\cite{snell2017prototypical} classify samples by their distance to class-mean prototypes in embedding space, and have been adapted to few-shot medical classification~\cite{ouyang2020self}.

PANDA departs from these lines of work in two key aspects. First, its prototypes are computed from the auxiliary-modality encoders rather than from the primary
(MRI) encoder, injecting auxiliary-modality class geometry into the MRI representation. 
Second, although Stage~1 trains the encoders jointly, the prototypes themselves are fixed as class means at the end of Stage~1 and are not learnable during Stage~2 fine-tuning; this avoids the circularity
that arises when a primary encoder is optimised against prototypes that are simultaneously updated from that same encoder. 
Empirically, making the prototypes learnable during MRI fine-tuning causes them to collapse onto the MRI encoder's class means, eroding their value as an independent supervisory signal.

\subsection{Survival Prediction from Pathology and Genomics}

Attention-based multiple instance learning (ABMIL)~\cite{ilse2018attention} is widely used for survival prediction from gigapixel whole-slide images (WSI), where patch-level attention aggregation yields interpretable slide-level representations.
Joint WSI\,+\,genomics models, including recent mixture-of-experts fusion~\cite{xiong2024mome}, can improve upon WSI-only survival prediction~\cite{chen2022porpoise,chen2021mcat}; however, they typically require RNA or mutation measurements at inference and are commonly trained only on the fully paired subset.
Recent pathology foundation models (e.g., UNI2~\cite{chen2024uni} and CONCH~\cite{lu2024conch}) provide high-quality patch embeddings that reduce labeled-data requirements for slide-level tasks.
In our TCGA validation, we use UNI2-h ABMIL features as the WSI representation and apply the same two-stage prototype framework with RNA-seq as the auxiliary modality, demonstrating transfer from neuroimaging to computational pathology without architectural modification.

% §2 Dataset
% ============================================================
% §2  DATASET
% Target venue: Springer Nature journal (sn-jnl)
% First draft, 2026-06-27
% ============================================================

\section{Dataset}
\label{sec:dataset}

% ── 2.1 ADNI AD/CN cohort ────────────────────────────────────────────────────
\subsection{ADNI AD/CN Cohort}
\label{sec:adni}

% ── Cohort composition figure (tikz) ─────────────────────────────────────────
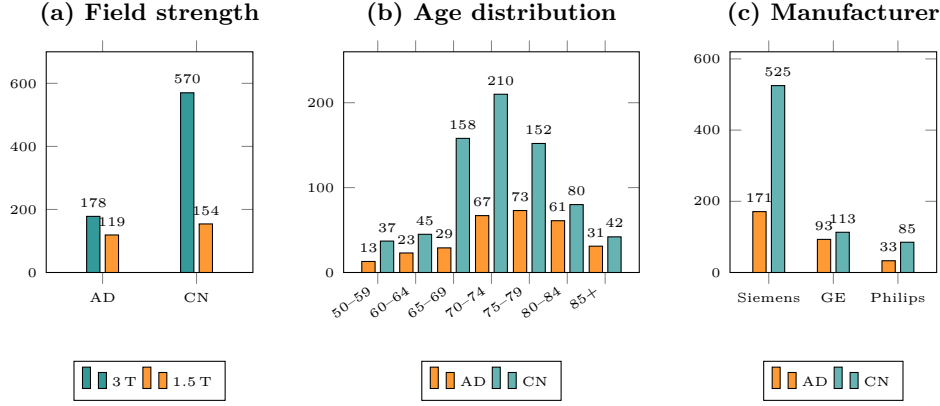
\begin{figure*}[t]
\centering
\begin{tikzpicture}
\begin{groupplot}[
  group style={group size=3 by 1, horizontal sep=1.2cm},
  height=4.5cm,
  ybar,
  enlarge x limits=0.4,
  nodes near coords,
  nodes near coords style={font=\tiny, anchor=south},
  tick label style={font=\tiny},
  label style={font=\tiny},
  title style={font=\small\bfseries},
  legend style={font=\tiny, at={(0.5,-0.40)}, anchor=north, legend columns=-1},
]

%% (a) Field strength by class
\nextgroupplot[
  width=0.33\textwidth,
  bar width=5pt,
  enlarge x limits=0.6,
  symbolic x coords={AD, CN},
  xtick=data,
  ymin=0, ymax=700,
  title={(a) Field strength},
]
\addplot[fill=teal!80]   coordinates {(AD,178) (CN,570)};
\addplot[fill=orange!80] coordinates {(AD,119) (CN,154)};
\legend{3\,T, 1.5\,T}

%% (b) Age distribution by class
\nextgroupplot[
  width=0.42\textwidth,
  bar width=5pt,
  enlarge x limits=0.15,
  symbolic x coords={50--59,60--64,65--69,70--74,75--79,80--84,85+},
  xtick=data,
  x tick label style={rotate=35, anchor=east, font=\tiny},
  ymin=0, ymax=260,
  title={(b) Age distribution},
]
\addplot[fill=orange!80] coordinates {
  (50--59,13)(60--64,23)(65--69,29)(70--74,67)
  (75--79,73)(80--84,61)(85+,31)};
\addplot[fill=teal!60] coordinates {
  (50--59,37)(60--64,45)(65--69,158)(70--74,210)
  (75--79,152)(80--84,80)(85+,42)};
\legend{AD, CN}

%% (c) Manufacturer by class
\nextgroupplot[
  width=0.33\textwidth,
  bar width=5pt,
  enlarge x limits=0.3,
  symbolic x coords={Siemens, GE, Philips},
  xtick=data,
  ymin=0, ymax=620,
  title={(c) Manufacturer},
]
\addplot[fill=orange!80] coordinates {(Siemens,171)(GE,93)(Philips,33)};
\addplot[fill=teal!60]   coordinates {(Siemens,525)(GE,113)(Philips,85)};
\legend{AD, CN}

\end{groupplot}
\end{tikzpicture}
\caption{Cohort composition of the 1{,}021-subject ADNI AD/CN dataset.
  \textbf{(a)}~Class distribution by field strength (AD: 297, CN: 724);
  AD subjects are proportionally more represented at 1.5\,T (40.1\%) than
  CN (21.2\%).
  \textbf{(b)}~Age distribution in 5-year bins; AD subjects are on average
  2.6 years older ($75.6\pm 7.8$ vs.\ $73.0\pm 7.3$ years).
  \textbf{(c)}~Manufacturer distribution; Siemens dominates both groups.}
\label{fig:dataset_bars}
\end{figure*}

Data were obtained from the Alzheimer's Disease Neuroimaging Initiative (ADNI)~\cite{ADNIdataset}, a longitudinal multi-site study that acquires structural MRI, PET imaging, and standardised neuropsychological assessments across the diagnostic spectrum.

For binary AD/CN classification, we constructed a cohort of 1{,}021 subjects (AD: 297, CN: 724) drawn from ADNI-1/GO, ADNI-2, and ADNI-3. To avoid subject-level leakage, we selected one T1-weighted MRI per subject (earliest available visit; highest-quality acquisition) and excluded multi-channel or anomalous scans. We then created a fixed split, stratified by diagnosis and scanner field strength prior to any modelling:
\begin{itemize}
  \item Training/validation: $n=844$ (AD: 249, CN: 595).
  \item Held-out test: $n=177$ (AD: 48, CN: 129).
\end{itemize}

Scanner heterogeneity is a defining feature of ADNI: 273 scans (26.7\%) were acquired at 1.5\,T and 748 (73.3\%) at 3\,T (test set: 61 at 1.5\,T, 116 at 3\,T). Throughout this work, 1.5\,T is defined as \texttt{protocol\_field\_strength}\,$\leq 2.0$ and 3\,T as \texttt{protocol\_field\_strength}\,$> 2.0$. Importantly, field strength is class-imbalanced (AD: 40.1\% at 1.5\,T vs. CN: 21.3\%; Fig.~\ref{fig:dataset_bars}\textbf{a}), which can act as a systematic confound and motivates the scanner-stratified analyses in Section~\ref{sec:scanner}.

Mean age at scan is $75.6 \pm 7.8$ years for AD and $73.0 \pm 7.3$ years for CN (two-sample $t$-test: $t = 4.55$, $p < 0.001$). The cohort is 53.7\% female (548/1{,}021); the AD group is slightly male-dominated (M: 165, F: 132; 55.6\% male), whereas the CN group is predominantly female (F: 416, M: 308; 57.5\% female), consistent with prior ADNI analyses~\cite{ADNIdataset}. The overall class distribution is AD: 29.1\% and CN: 70.9\%. Scanner manufacturer distribution is Siemens 696 (68.2\%), GE 206 (20.2\%), and Philips 118 (11.6\%). Fig.~\ref{fig:dataset_bars} summarises these distributions.

% ── 2.2 Auxiliary modalities ─────────────────────────────────────────────────
\subsection{Auxiliary Modalities}
\label{sec:aux_modalities}

\textbf{Clinical tabular scores.} Table~\ref{tab:tabular_miss} summarises the availability of the clinical tabular variables in the 844-subject training/validation cohort. Standardised neuropsychological and demographic variables were extracted from the ADNI repository, and we retained three key features: Mini-Mental State Examination (MMSE), Clinical Dementia Rating global score (CDR), and Functional Activities Questionnaire total score (FAQ). Cognitive and functional assessments exhibit substantial missingness, with each feature available for only about half of subjects (FAQ: 51.5\%, MMSE: 45.1\%, CDR: 44.8\% available; Table~\ref{tab:tabular_miss}), consistent with variable assessment schedules across ADNI phases.

We define the fully paired subset $\mathcal{D}_\mathrm{paired}$ as the set of subjects for whom all three tabular features are simultaneously observed. This yields $|\mathcal{D}_\mathrm{paired}|=378$ training subjects (AD: 170, CN: 208), corresponding to a tabular pairing rate of $r_\mathrm{tab}=44.8\%$. All tabular-prototype experiments use $\mathcal{D}_\mathrm{paired}$ as the Stage~1 anchor pool; feature scaling is fit independently within each training fold to prevent information leakage.

% ---- TABLE: Tabular feature availability ----
\begin{table}[t]
\centering
\caption{Tabular feature availability in the 844-subject training/validation
  cohort (AD: 249, CN: 595). The fully paired subset $\mathcal{D}_\mathrm{paired}$
  comprises subjects for whom all three features are simultaneously available.}
\label{tab:tabular_miss}
\setlength{\tabcolsep}{5pt}
\renewcommand{\arraystretch}{1.15}
\begin{tabular*}{\linewidth}{@{\extracolsep{\fill}}lrrr}
\toprule
\textbf{Feature} & \textbf{Available} & \textbf{Missing} & \textbf{Available (\%)} \\
\midrule
FAQ total score                      & 435 & 409 & 51.5 \\
MMSE score                           & 381 & 463 & 45.1 \\
CDR global score                     & 378 & 466 & 44.8 \\
\midrule
$\mathcal{D}_\mathrm{paired}$ (all 3 features) & 378 & 466 & 44.8 \\
\bottomrule
\end{tabular*}
\end{table}

\noindent
\textbf{FDG-PET volumes.} FDG-PET scans were obtained from the ADNI repository for a subset of the AD/CN cohort. Following DICOM-to-NIfTI conversion, 248 FDG-PET volumes were available, spanning both static and dynamic acquisitions under standard ADNI protocols. Within each training fold, approximately 158 subjects (AD: $\approx$70, CN: $\approx$88) were matched to a corresponding MRI scan; we denote this PET-paired subset by $\mathcal{S}_\mathrm{PET}$, yielding a pairing rate of $r_\mathrm{PET} \approx 18.7\%$. PET availability is concentrated in later ADNI phases and is therefore predominantly associated with 3\,T imaging, which makes PET anchors relatively sparse for the 1.5\,T scans where scanner-induced bias is most pronounced.

\noindent
\textbf{Handwriting kinematics (DARWIN).} The DARWIN (Diagnosis AlzheimeR WIth haNdwriting) dataset~\cite{cilia2022diagnosing} is a publicly available handwriting kinematics benchmark comprising 174 participants (AD and healthy controls) acquired using a digitising tablet.
Participants completed 25 standardised handwriting and drawing tasks. For each task, 18 kinematic descriptors (e.g., velocity, pressure, and pen-up/pen-down ratio) were extracted, yielding a $25\times 18=450$-dimensional feature vector per subject.
DARWIN shares no subjects, sites, or acquisition hardware with the ADNI cohort; consequently, the handwriting pairing rate is $r_\mathrm{HW}=0$. DARWIN class prototypes are computed on the DARWIN cohort and transferred directly into Stage~2 of ADNI training without any shared subjects.

% ── 2.3 MCI cohort (held-out) ────────────────────────────────────────────────
\subsection{MCI Cohort for Zero-Shot Severity Evaluation}
\label{sec:mci_data}

We additionally define an independent cohort of 147 subjects with mild cognitive impairment (MCI) and resolvable MRI in the same image-quality-filtered pool as the AD/CN test set (EMCI: 49, MCI: 79, LMCI: 19). This cohort is withheld from all training stages, and no MCI scans, labels, or clinical scores are used for model fitting. Because these subjects are drawn from ADNI using the same sites and hardware as the AD/CN cohort, they are subject to the same 1.5\,T/3\,T field-strength imbalance.

We use this cohort exclusively for post hoc evaluation of whether the learned MRI representations encode an ordinal disease-severity gradient beyond binary AD/CN separation. Severity proxies include MMSE, CDR global score, FAQ total score, and NPI total score. Ordinal consistency is assessed using Kruskal--Wallis tests and one-sided Mann--Whitney tests across EMCI/MCI/LMCI, together with a monotonicity check against the CN/AD reference groups (Section~\ref{sec:mci}).

% ── 2.4 TCGA-Lung ────────────────────────────────────────────────────────────
\subsection{TCGA-Lung: Cross-Domain Validation}
\label{sec:tcga_data}

To assess whether the proposed prototype-alignment mechanism transfers beyond neuroimaging, we evaluate PANDA on a cross-domain histopathology survival prediction task using TCGA-Lung (TCGA-LUAD\,+\,TCGA-LUSC).
This setting instantiates the same training constraint as ADNI, but with whole-slide imaging (WSI) features as the primary modality and bulk RNA-seq as the auxiliary modality.

\noindent
\textbf{Modalities and availability.} Diagnostic H\&E WSI constitute the primary input, and bulk RNA-seq (downloaded from the GDC portal) serves as the auxiliary modality. We consider two endpoints: (i) binary 2-year overall survival ($n=594$) and (ii) Cox proportional-hazards survival modelling ($n=853$). WSI is available for all subjects, whereas RNA-seq is available for 591/594 (99.5\%) and 849/853 (99.5\%), respectively, corresponding to a near-complete pairing regime.

%Implementation details for feature extraction, preprocessing, task definitions, and data splits are provided in Section~\ref{sec:tcga_setup}.
%evaluation metrics are given in Section~\ref{sec:setup}.

% §3 Method / §4 Experiments / §5 Discussion
% ============================================================
% §3  METHOD  /  §4  EXPERIMENTS  /  §5  DISCUSSION
% Target venue: Springer Nature journal (sn-jnl)
% First draft, 2026-06-27
% ============================================================

\section{Method}
\label{sec:method}

\begin{figure*}[t!]
\centering
\includegraphics[width=\textwidth]{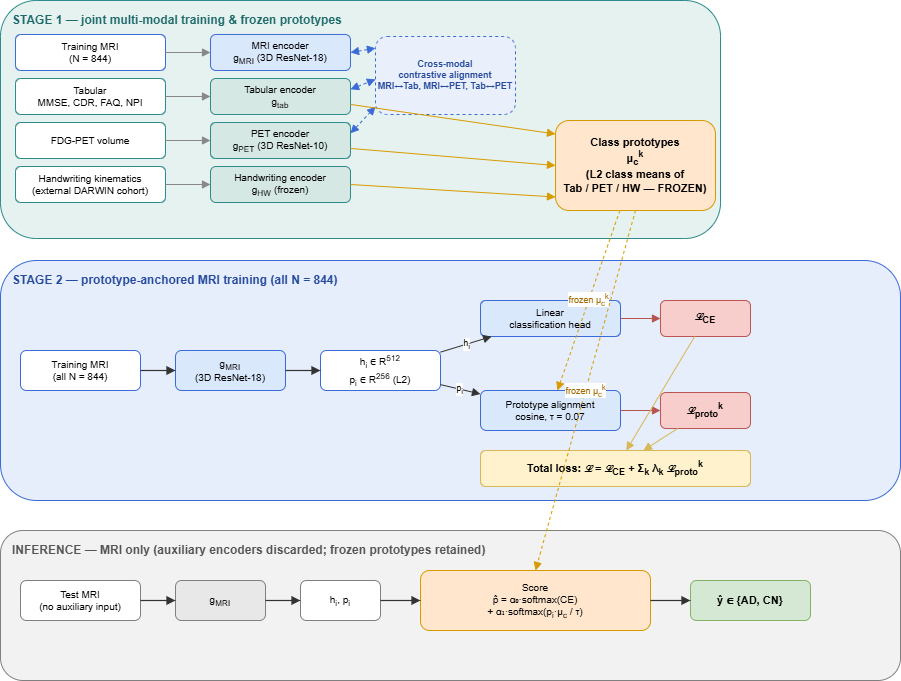}
\caption{Overview of the two-stage prototype-anchored alignment framework.
\textbf{Stage~1} jointly trains the MRI encoder together with the auxiliary
encoders (tabular, PET, external handwriting) under a single combined objective
(per-modality cross-entropy and pairwise cross-modal contrastive losses), then
computes frozen class prototypes $\boldsymbol{\mu}_c^{(k)}$ from the auxiliary
embeddings. \textbf{Stage~2} trains the MRI encoder
$g_\mathrm{MRI}$ on \emph{all} $N=844$ subjects with the combined objective
$\mathcal{L}=\mathcal{L}_\mathrm{CE}+\sum_k\lambda_k\mathcal{L}^{(k)}_\mathrm{proto}$
($\tau=0.07$), pulling the normalised MRI projection $\mathbf{p}_i$ toward the
frozen prototypes. At \textbf{inference} all auxiliary encoders are discarded
and the class score is the convex blend of the linear-head and prototype-cosine
softmaxes (Eq.~\ref{eq:infer_blend}); only MRI is required.}
\label{fig:overview}
\end{figure*}

A standard baseline in multimodal learning assumes per-subject pairing and improves performance either by fusing modalities in a joint model or by enforcing cross-modal alignment with paired samples (e.g., contrastive objectives). These designs typically require auxiliary modalities to be present for a substantial subset of subjects and, in many cases, at deployment.

In our setting, auxiliary measurements are heterogeneous and can be partially paired, sparsely paired, or available only in an external cohort with no subject overlap, while the deployed model must operate on the primary modality alone. PANDA addresses this mismatch by using auxiliary data only to estimate class-level anchors. In Stage~1, auxiliary-modality encoders are used to compute class prototypes, which are then frozen; in Stage~2, the primary encoder is trained to align its representations to these fixed prototypes for all subjects. Fig.~\ref{fig:overview} summarises the overall procedure.

\subsection{Framework}
   
\paragraph{Problem Formulation} Let $\mathcal{D} = \{(\mathbf{x}_i, y_i)\}_{i=1}^{N}$ denote a dataset of
$N$ subjects, where $\mathbf{x}_i$ is a 3-D T1-weighted MRI volume and $y_i \in \{0, 1\}$ is the binary AD/CN label.
Every subject has an MRI scan; the primary classifier $f: \mathbf{x} \mapsto \hat{y}$ must operate on MRI alone at inference.

In addition, $K$ auxiliary modalities are available, each observed for a
\emph{subset} of subjects.
For modality $k \in \{1, \ldots, K\}$, let $\mathcal{S}_k \subseteq \{1,\ldots,N\}$
denote the paired subset, with pairing rate $r_k = |\mathcal{S}_k| / N$.
In our study, $K = 3$ auxiliary modalities are considered:
\begin{itemize}
  \item \textbf{Tabular clinical scores} (MMSE, CDR, FAQ): $r_\mathrm{tab} \approx 0.45$
        ($n_\mathrm{tab} = 378$ of 844 training subjects).
  \item \textbf{FDG-PET volumes}: $r_\mathrm{PET} \approx 0.19$
        ($n_\mathrm{PET} \approx 158$ per training fold).
  \item \textbf{Handwriting kinematics} (DARWIN dataset~\cite{cilia2022diagnosing}):
        $r_\mathrm{HW} = 0$, with no subject overlap with the ADNI cohort.
        Prototypes are derived from the external cohort and transferred directly.
\end{itemize}

The goal is to train $f$ such that (i) it exploits all available auxiliary supervision at training time, (ii) it requires \emph{no} auxiliary input at inference, and (iii) its learned representation is robust to scanner field-strength variation (1.5\,T vs.\ 3\,T).

% ── Architecture overview ────────────────────────────────────────────────────
\paragraph{Encoder Architecture}
\label{sec:arch}

\textbf{MRI backbone.}
The primary encoder $g_\mathrm{MRI}$ is a 3-D ResNet-18 initialised from MedicalNet~\cite{chen2019med3d}, pretrained on a large corpus of volumetric segmentation tasks.
It maps each $128^3$-voxel input to a 512-dimensional feature vector
$\mathbf{h}_i \in \mathbb{R}^{512}$.
A two-layer MLP projection head $\phi_\mathrm{MRI}$ reduces $\mathbf{h}_i$
to a 256-dimensional embedding $\mathbf{p}_i$, which is $\ell_2$-normalised
to the unit hypersphere:
\begin{equation}
  \mathbf{p}_i = \frac{\phi_\mathrm{MRI}(\mathbf{h}_i)}
                      {\|\phi_\mathrm{MRI}(\mathbf{h}_i)\|_2}.
  \label{eq:mri_proj}
\end{equation}
A small constant is added to norm denominators (here and in the prototype normalisation) for numerical stability. The classification head is a single linear layer applied to the 512-dimensional backbone features $\mathbf{h}_i$ (before projection), so that classification and representation alignment are trained with separate parameter paths.

\noindent
\textbf{Auxiliary encoders.}
Each auxiliary modality $k$ has its own encoder $g_k$ that maps its
modality-specific input to a 256-dimensional $\ell_2$-normalised embedding
$\mathbf{z}_i^{(k)}$ in the \emph{same} shared hypersphere as $\mathbf{p}_i$:
\begin{itemize}
  \item \textit{Tabular encoder}: a three-layer MLP ($d_\mathrm{tab} \to 128 \to 256$) with LayerNorm and dropout (0.3) after each hidden layer.
    Subjects with missing values in any tabular input are excluded from tabular-encoder training and from the prototype anchor pool, ensuring that prototypes are computed from complete feature vectors.
  \item \textit{PET encoder}: a 3-D ResNet-10 (a lightweight MedicalNet variant~\cite{chen2019med3d}) applied to $128^3$-voxel intensity-normalised PET volumes, followed by a two-layer MLP projection head matching the MRI head.
  \item \textit{Handwriting encoder}: a two-layer MLP ($450 \to 256 \to 256$) applied to the concatenated DARWIN handwriting kinematics vector (25 tasks $\times$ 18 features).
    This encoder is trained exclusively on the external DARWIN cohort; its outputs are used only to compute class prototypes, and its parameters remain frozen throughout Stage~2.
\end{itemize}

% ── Class prototypes ─────────────────────────────────────────────────────────
\paragraph{Class Prototype Construction}
\label{sec:prototypes}

For each auxiliary modality $k$ and class $c \in \{0, 1\}$, the class prototype
$\boldsymbol{\mu}_c^{(k)}$ is the $\ell_2$-normalised mean embedding over all
paired training subjects of that class:
\begin{equation}
  \boldsymbol{\mu}_{c}^{(k)} =
    \operatorname{Normalize}\!\left(
      \frac{1}{|\mathcal{S}_k^c|}
      \sum_{i \in \mathcal{S}_k^c} \mathbf{z}_{i}^{(k)}
    \right),
    \quad \mathcal{S}_k^c = \{i \in \mathcal{S}_k : y_i = c\}.
  \label{eq:prototype}
\end{equation}
Prototypes are computed \emph{once} from the jointly-trained auxiliary encoders at the end of Stage~1 and remain \textbf{frozen} for the remainder of training.
Freezing prevents prototype drift and ensures that the MRI encoder optimises toward stable geometric targets.

For the handwriting modality ($r_\mathrm{HW}=0$), the paired subset $\mathcal{S}_\mathrm{HW}$ contains no ADNI subjects. We therefore compute handwriting class prototypes on the external DARWIN cohort and transfer them directly as $\{\boldsymbol{\mu}_0^{\mathrm{HW}},\boldsymbol{\mu}_1^{\mathrm{HW}}\}$.
This \emph{external prototype transfer} regime evaluates whether class-level geometry learned in an entirely separate cohort can act as an effective anchor for MRI representation learning.
Moreover, it serves as a control against the hypothesis that prototype alignment requires subject-level correspondence between modalities: any observed benefit in this regime must arise from the transferred class separation in prototype space rather than from per-subject cross-modal pairing.

% ── Two-stage training ───────────────────────────────────────────────────────
\subsection{Training and Inference scheme}
\label{sec:training}

\noindent
\textbf{Stage 1: joint multi-modal training and frozen prototypes.}
Stage~1 jointly trains the MRI encoder and all auxiliary encoders using a single optimiser and a unified objective. The loss comprises (i) per-modality cross-entropy terms for modalities with AD/CN labels and (ii) supervised cross-modal contrastive terms over the available paired subsets (MRI--tabular, MRI--PET, and tabular--PET), yielding a shared $\ell_2$-normalised embedding space. The handwriting encoder is trained on the external DARWIN cohort and contributes only class-level prototypes (no subject overlap with ADNI).
After Stage~1, all auxiliary encoders are frozen and class prototypes $\{\boldsymbol{\mu}_c^{(k)}\}$ are computed once via \eqref{eq:prototype} and held fixed for Stage~2.

\noindent
\textbf{Stage 2: prototype-anchored MRI training.}
The MRI encoder is then trained on the \emph{full} dataset $\mathcal{D}$,
regardless of auxiliary modality availability, with the combined loss
\begin{equation}
  \mathcal{L} = \mathcal{L}_{\mathrm{CE}} + \sum_{k=1}^{K} \lambda_k
                \mathcal{L}^{(k)}_{\mathrm{proto}},
  \label{eq:total_loss}
\end{equation}
where $\mathcal{L}_{\mathrm{CE}}$ is the cross-entropy loss on AD/CN logits
computed from backbone features $\mathbf{h}_i$, each $\lambda_k$ weights
modality $k$, and each $\mathcal{L}^{(k)}_{\mathrm{proto}}$ is a temperature-scaled prototype
cross-entropy~\cite{snell2017prototypical} that pulls the normalised MRI
projection $\mathbf{p}_i$ toward the frozen prototype of its correct class:
\begin{equation}
  \mathcal{L}^{(k)}_{\mathrm{proto}} =
    -\frac{1}{|\mathcal{B}|}\sum_{i \in \mathcal{B}}
      \log \frac{
        \exp\!\bigl(\mathbf{p}_i \cdot \boldsymbol{\mu}^{(k)}_{y_i} / \tau\bigr)
      }{
        \sum_{c} \exp\!\bigl(\mathbf{p}_i \cdot \boldsymbol{\mu}^{(k)}_{c} / \tau\bigr)
      },
  \label{eq:proto_loss}
\end{equation}
with $\mathcal{B}$ a training mini-batch, $\tau$ the temperature
parameter~\cite{radford2021learning}, and $c \in \{0, 1\}$ ranging over both
classes.
The defining property of this objective is that the same prototype
$\boldsymbol{\mu}^{(k)}_{y_i}$ is used for \emph{every} subject $i$, whether or
not $i \in \mathcal{S}_k$: paired and unpaired subjects receive identical
geometric targets for a given modality, so no subject is excluded from
alignment supervision. This is what allows the MRI encoder to learn from the
entire training cohort even though the auxiliary signal is defined on the paired
subset alone. 
%Optimisation hyperparameters, the temperature $\tau$, and the loss weights $\lambda_k$ are given in Section~\ref{sec:setup}.

% Inference
\paragraph{Inference}
\label{sec:inference}

At test time, all auxiliary encoders are discarded.
Classification of a new subject uses only the MRI backbone $g_\mathrm{MRI}$ and the linear classification head, incurring no additional computational cost
relative to the MRI-only baseline.
At inference the class score for a test subject is the weighted average of the linear-head softmax, whose logits are $\ell(\mathbf{h}_i)$, and the prototype cosine softmax,
\begin{equation}
\begin{aligned}
  \hat{p}(c\mid\mathbf{x}_i)
  &= \alpha_0\,\mathrm{softmax}_c\bigl(\ell(\mathbf{h}_i)\bigr)
   + \alpha_1\,\mathrm{softmax}_c\bigl(\textstyle\sum_{k\in\{\mathrm{Tab},\mathrm{PET},\mathrm{HW}\}} \mathbf{p}_i \cdot \boldsymbol{\mu}^{(k)}_c / \tau\bigr), \\
  &\quad \alpha_0,\alpha_1 \in [0,1], \qquad \alpha_0 + \alpha_1 = 1,
\end{aligned}
  \label{eq:infer_blend}
\end{equation}
with $\hat{y}=\arg\max_c \hat{p}(c\mid\mathbf{x}_i)$. The weight $(\alpha_0,\alpha_1)$ is fit on the validation fold (Implementation Details, Section~\ref{sec:implementation}); the reported
$p_\mathrm{AD}$ (including the zero-shot MCI analysis of
Section~\ref{sec:mci}) is $\hat{p}(\text{AD}\mid\mathbf{x}_i)$. When
$\alpha_1=0$ this reduces to the MRI-only linear classifier. The prototype-cosine term sums the raw cosine logits against all three frozen prototype sets ($k\in\{\mathrm{Tab},\mathrm{PET},\mathrm{HW}\}$) with equal weight and a single shared temperature $\tau=0.07$, followed by a single softmax; since the prototypes $\boldsymbol{\mu}^{(k)}_c$ are frozen constants, no auxiliary input is required at inference and the classifier remains MRI-only for every test subject.

\paragraph{Severity-Axis Extension}
\label{sec:severity_head}

The base framework is trained for binary CN/AD discrimination and does not explicitly enforce an ordinal severity structure (Section~\ref{sec:mci}). We therefore augment Stage~2 with a linear severity head $h_\mathrm{sev}: \mathbb{R}^{512} \to \mathbb{R}$ applied to the $512$-dimensional backbone features $\mathbf{h}_i$ (the pre-projection encoder output), so that the severity signal is read off features independent of the tabular/PET/hand-crafted-aligned projection space, and trained by masked Huber regression against graded clinical-severity composites computed on the CN/AD training pool only:
\begin{equation}
\mathcal{L}_\mathrm{sev} = \frac{1}{|\mathcal{S}|} \sum_{i \in \mathcal{S}} \mathrm{Huber}\big(h_\mathrm{sev}(\mathbf{h}_i),\, s_i\big),
\end{equation}
where $\mathcal{S}$ denotes CN/AD training subjects with complete coverage for the chosen composite and $s_i$ is the corresponding z-scored value.
We consider two composites: $s_\mathrm{orth}$ (mean of z-scored FAQ and NPI; $n=265$ of 844 CN/AD training subjects) and $s_\mathrm{diag}$ (mean of z-scored CDR and negated z-scored MMSE; $n=378$). Unless otherwise stated, we use $s_\mathrm{orth}$ because FAQ/NPI reflect functional and neuropsychiatric status rather than directly restating the diagnostic criteria used to assign CN/AD labels. In the CN/AD training pool, FAQ is observed for $435$ subjects and NPI for $265$; because every NPI-observed subject also has FAQ, NPI is the limiting factor and the complete-coverage set for $s_\mathrm{orth}$ is exactly these $n=265$ subjects. Each component is z-scored independently over all CN/AD subjects for which it is observed (FAQ over $435$, NPI over $265$) before averaging. Missing values are handled by exclusion rather than imputation: the masked Huber loss requires both raw FAQ and NPI to be present, so only the $265$ complete-coverage subjects contribute gradients ($s_\mathrm{diag}$ is defined analogously, requiring both CDR and MMSE, $n=378$). This complete-coverage subset is also field-strength--skewed (208 subjects at $\leq2.0$\,T vs.\ 57 at $>2.0$\,T), so the learned severity axis is confounded by lower-field scanner characteristics; we therefore treat it as an exploratory readout and interpret absolute severity values with corresponding caution (see also Section~\ref{sec:limitations}).

The severity head is optimised jointly with the Stage~2 objective,
$\mathcal{L}_\mathrm{CE} + 0.3(\mathcal{L}_\mathrm{tab}+\mathcal{L}_\mathrm{PET}+\mathcal{L}_\mathrm{HW}) + \lambda_\mathrm{sev}\mathcal{L}_\mathrm{sev}$ with $\lambda_\mathrm{sev}=0.3$; equivalently, this is the general Stage-2 objective of \eqref{eq:total_loss} with $\lambda_k=0.3$ for all prototype terms, augmented by the severity term.
Stage~1 is reused from the frozen joint checkpoint (i.e., not retrained), isolating the effect of the severity head from changes in the Stage~1 alignment geometry.
MCI subjects contribute no training gradients at any stage: the composite is defined and standardised using CN/AD subjects only, and the head is evaluated on MCI in a zero-shot manner.

% ── Information-theoretic motivation ─────────────────────────────────────────
\subsection{Information-Theoretic Motivation}
\label{sec:infotheory}

We provide a brief information-theoretic motivation for prototype anchoring and for the empirical observation that reduced pairing need not degrade performance. These arguments are heuristic, intended to build intuition rather than to constitute rigorous proofs, and rely on simplifying assumptions that we state where used. Let $Y\in\{0,1\}$ denote the class label, $\mathbf{z}^{(k)}$ the Stage~1 auxiliary embedding for modality $k$, and $\mathbf{p}$ the $\ell_2$-normalised MRI projection in \eqref{eq:mri_proj}.

\paragraph{Alignment transfers class information.}
The Stage~1 cross-modal InfoNCE objective provides a variational lower bound on the mutual information between paired MRI and auxiliary embeddings,
$I(\mathbf{p};\mathbf{z}^{(k)}) \ge \log N_\mathcal{B} - \mathcal{L}_{\mathrm{NCE}}$~\cite{poole2019variational}. This bound pertains to the InfoNCE term alone; the full Stage~1 objective, which also includes per-modality cross-entropy terms, does not directly inherit it. The class prototype $\boldsymbol{\mu}^{(k)}_c = \operatorname{Normalize}\big(\mathbb{E}[\mathbf{z}^{(k)} \mid Y=c]\big)$ preserves class-discriminative directions of the auxiliary embedding. Minimising the prototype cross-entropy in \eqref{eq:proto_loss} therefore encourages $\mathbf{p}$ to align with the auxiliary-induced class geometry~\cite{boudiaf2020unifying}. Because the target prototypes are fixed and class-level, this supervisory signal is available to all subjects during Stage~2, including those without paired auxiliary measurements.

\paragraph{Prototype separation as an anchor-quality proxy.}
Consider the cosine score $s_c = \mathbf{p} \cdot \boldsymbol{\mu}^{(k)}_c / \tau$ under a two-class Gaussian-channel approximation with equal class-conditional variances. Then $I(s;Y)$ is monotone in the deflection coefficient
$J = \lVert \boldsymbol{\mu}^{(k)}_1 - \boldsymbol{\mu}^{(k)}_0 \rVert^2 / (\sigma_0^2 + \sigma_1^2)$ (assuming $\sigma_0^2+\sigma_1^2>0$), so larger inter-prototype separation yields a stronger anchor. Since prototypes are finite-sample estimates of class means, we have
$\mathbb{E}\lVert\hat{\boldsymbol{\mu}}_1-\hat{\boldsymbol{\mu}}_0\rVert^2 = \lVert\boldsymbol{\mu}_1-\boldsymbol{\mu}_0\rVert^2 + \operatorname{tr}\,\operatorname{Var}(\hat{\boldsymbol{\mu}}_1-\hat{\boldsymbol{\mu}}_0)$,
which implies that subsampling can increase empirical separation while also increasing estimator variance. This identity assumes unnormalised class means; for the $\ell_2$-normalised prototypes used here it holds only approximately, and the bias term is also affected by the normalisation. Consequently, moderate reductions in pairing can leave the empirical anchor quality unchanged (or slightly improved) until the class-mean estimates become too noisy (empirically, around $\approx 40$ paired AD subjects per fold; Section~\ref{sec:pair_rate}). Importantly, this is an estimator effect on empirical prototypes rather than an increase in the population mutual information $I(\mathbf{z}^{(k)};Y)$. We test this prediction by measuring inter-prototype separation as a function of pairing rate in Section~\ref{sec:pair_rate}.
\section{Experimental Setup}
\label{sec:setup}

% ── 4.2 Implementation ───────────────────────────────────────────────────────
\paragraph{Implementation Details}
\label{sec:implementation}

MRI volumes were resampled to 1\,mm isotropic in RAS orientation and cropped to $128^3$ voxels using MONAI~\cite{cardoso2022monai}.
Intensity normalisation (zero mean, unit variance over non-background voxels) and training-time augmentation (random affine: $\pm 10^\circ$,
$\pm 10\%$ scale, $\pm 10$\,mm translation) were applied.
FDG-PET volumes were preprocessed to $128^3$ voxels at 2\,mm isotropic resolution with 99th-percentile intensity normalisation. 
All experiments use three random but fixed seeds under 5-fold stratified cross-validation on the training set; the test set is evaluated once per
seed using the best validation-AUC checkpoint.
Reported metrics are mean\,$\pm$\,SD across the three seeds.

{Stage 1} uses a single AdamW optimiser (weight decay $10^{-4}$) with fixed learning rates and no scheduler: MRI parameters (\texttt{layer3}/\texttt{layer4} and the projection head) use lr $=10^{-4}$ and
the auxiliary encoders (tabular, PET, handwriting) use lr $=5\times10^{-5}$, trained for 50 epochs (handwriting: 100). Tabular feature scaling is fit on the
training fold only. We set $\tau=0.07$ and $\lambda_\mathrm{tab}=0.5$, $\lambda_\mathrm{PET}=0.3$, $\lambda_\mathrm{HW}=0.5$ on validation AUC at 100\% pairing, held fixed across pairing-rate ablations.
{Stage 2} then fine-tunes the MRI encoder alone, also with AdamW but at a lower backbone rate ($10^{-5}$; projection head $10^{-4}$), for up to 40 epochs
with early stopping (patience $=10$) on validation AUC and ReduceLROnPlateau (factor $=0.5$, patience $=3$). All runs execute on 2$\times$A100 GPUs.
The inference blend weights $(\alpha_0,\alpha_1)$ in \eqref{eq:infer_blend} are fit independently for each Stage-2 fold (and seed) on that fold's held-out validation split---the same split used for checkpoint selection, with no separate nested cross-validation---by minimising the negative log-likelihood of the blended probability against the validation labels. The weights are not grid-searched: $(\alpha_0,\alpha_1)=\mathrm{softmax}(\boldsymbol{\theta})$ is parameterised by a single learnable $2$-vector $\boldsymbol{\theta}$ (initialised at zero) and optimised with Adam (learning rate $10^{-3}$) for $200$ full-batch epochs, retaining the best-validation-loss epoch. At test time the five folds are combined by soft-voting (averaging the blended $\hat{p}$). Across the $15$ fold/seed fits the objective is bimodal rather than a single stable optimum: roughly half the folds converge near $(0.40,0.60)$ and the remainder to the mirror configuration $(0.59,0.41)$ that up-weights the linear head, giving a mean of $(\alpha_0,\alpha_1)=(0.487\pm0.091,\,0.513\pm0.091)$ across folds. The soft-voting ensemble averages over this per-fold variation, and the near-equal mean indicates the two heads contribute comparably overall.

% ── 4.3 Baselines ────────────────────────────────────────────────────────────
\paragraph{Baselines}
\label{sec:baselines}

Unless otherwise stated, ADNI baselines are trained on the full 844-subject training/validation set and evaluated on the held-out test set ($n=177$) with MRI-only inputs at inference. Auxiliary modalities, when used, are provided during training only. We compare against 13 ADNI baselines and 2 TCGA baselines spanning (i) standard multimodal fusion, (ii) missing-modality training strategies, (iii) teacher--student distillation, and (iv) recent published methods. For each comparator, Appendix~\ref{app:baselines} (Table~\ref{tab:baselines}) specifies the training-time auxiliary inputs, inference-time inputs, and any protocol adaptations required for fair comparison.

\paragraph{Pairing-Rate Ablation Protocol}
\label{sec:pairing_rate}

To quantify sensitivity to reduced auxiliary pairing, we subsample the paired pools $\mathcal{S}_k$ prior to Stage~2. For the joint Tab+PET anchor we consider fractions $r \in \{1.0, 0.75, 0.5, 0.25, 0.15, 0.10, 0.05\}$, and for the tabular-only anchor we use $r \in \{1.0, 0.5, 0.25\}$. Subsampling is performed independently per modality and stratified by class to preserve class proportions within each prototype pool.
For each fraction, we reuse the jointly trained Stage~1 encoder weights and recompute only the prototype vectors $\boldsymbol{\mu}_c^{(k)}$ on the subsampled pools. Stage~2 training (MRI encoder on all $N=844$ subjects) and inference are otherwise unchanged. This protocol isolates the effect of prototype estimation quality from changes in the size of the MRI training set.

\paragraph{Cross-Domain Evaluation Setup}
\label{sec:tcga_setup}

To assess domain generality, we apply PANDA, unchanged, to a histopathology survival task on TCGA-Lung (LUAD\,+\,LUSC).
Here the primary modality is whole-slide images (WSI): diagnostic H\&E slides are tiled at 20$\times$ magnification, encoded with the UNI2-h pathology foundation model~\cite{chen2024uni}, and aggregated per slide by attention-based multiple-instance learning (ABMIL) into a 1536-dimensional bag embedding. The auxiliary modality is bulk RNA-seq ($\approx$99\% pairing): raw counts are reduced to the top-2000 most variable genes, log1p-transformed, and z-scored per gene on the training fold.
We evaluate two survival endpoints. For binary 2-year overall survival, subjects are labelled as deceased within 730\,days ($1$) or surviving beyond 730\,days ($0$); censored cases with follow-up $\leq 730$\,days are excluded. This yields $n=594$ subjects (test $=119$) with a label-stratified split (\texttt{random\_state}$=0$). For Cox proportional-hazards modelling, we include all subjects with valid follow-up and treat censored cases through the partial likelihood, yielding $n=853$ (test $=171$) with stratification by event indicator.

All TCGA experiments use the same three random seeds as the ADNI setup. To probe sensitivity to reduced auxiliary coverage, we subsample the RNA-paired pool at $r_\mathrm{RNA} \in \{0.25, 0.50, 1.00\}$. We report AUC, balanced accuracy, and macro-F1 for the binary endpoint and Harrell's C-index for the Cox model; all other architectural and optimisation settings are unchanged. This cross-domain evaluation tests whether the proposed prototype-alignment mechanism extends beyond neuroimaging and beyond sparsely paired auxiliary modalities.

\section{Results}
\label{sec:results}

\subsection{Main Results}
\label{sec:main_results}

% ---- TABLE: Main results ----
\begin{sidewaystable}
\centering
\footnotesize
\caption{Held-out test performance. Training-set size (number of subjects) is
given in parentheses after each method: the top block trains on the full
844-subject pool (exploiting unpaired subjects), the bottom block on the 378
tab-complete subjects only. \emph{Training} lists the modalities available
during training. Unless noted, evaluation is on the held-out test set
($n = 177$; AD $= 48$, CN $= 129$) with MRI-only inputs at inference.
Mean\,$\pm$\,SD across 3 seeds. Bold indicates the best value per column.
Bootstrap 95\,\% CIs for primary comparisons vs.\ MRI-only are reported in the
text (10\,000 iterations on seed-averaged probabilities).
Significance of the AUC gain vs.\ MRI-only (bootstrap with Holm--Bonferroni
correction): $^{*}p<0.05$, $^{**}p<0.01$, $^{***}p<0.001$.
$^{\dagger}$Suk et al.\ is evaluated on the paired test subset ($n = 75$)
rather than the full $n = 177$ set, as its late-fusion
design requires both modalities at inference; it is therefore not directly
comparable to the other rows. The final block lists independently-published
architectures with different backbones, run on our protocol for reference and
excluded from the per-column bold.}
\label{tab:main_results}
\setlength{\tabcolsep}{4pt}
\renewcommand{\arraystretch}{1.2}
\begin{tabular}{@{}llcccccc@{}}
\toprule
\textbf{Method (Training size)}
  & \textbf{Training}
  & \makecell{\textbf{AUC} $\uparrow$}
  & \makecell{\textbf{BalAcc} $\uparrow$}
  & \makecell{\textbf{MacroF1} $\uparrow$}
  & \makecell{\textbf{F1-AD} $\uparrow$}
  & \makecell{\textbf{Rec-AD} $\uparrow$}
  & \makecell{\textbf{Rec-CN} $\uparrow$} \\
\midrule
% ---- Top block: full 844-subject training pool ----
PANDA (844)
  & MRI+Tab
  & $0.790_{\pm.012}$
  & $0.692_{\pm.021}$
  & $0.689_{\pm.019}$
  & $0.550_{\pm.031}$
  & $0.563_{\pm.045}$
  & $0.822_{\pm.011}$ \\
PANDA (844)
  & MRI+Tab+PET
  & $0.851_{\pm.010}$
  & $0.763_{\pm.036}$
  & $0.733_{\pm.037}$
  & $0.638_{\pm.046}$
  & $\mathbf{0.750_{\pm.074}}$
  & $0.775_{\pm.055}$ \\
\textbf{PANDA} (844)
  & MRI+Tab+PET+HW
  & $\mathbf{0.868_{\pm.020}}^{*}$
  & $\mathbf{0.772_{\pm.041}}$
  & $\mathbf{0.767_{\pm.033}}$
  & $\mathbf{0.662_{\pm.054}}$
  & $0.681_{\pm.087}$
  & $0.863_{\pm.007}$ \\
\addlinespace
MRI-only Baseline (844)
  & MRI
  & $0.789_{\pm.016}$
  & $0.691_{\pm.019}$
  & $0.670_{\pm.009}$
  & $0.546_{\pm.026}$
  & $0.632_{\pm.080}$
  & $0.749_{\pm.045}$ \\
Knowledge distillation~\cite{hu2020knowledge} (844)
  & MRI+Tab+PET
  & $0.777_{\pm.016}$
  & $0.682_{\pm.008}$
  & $0.669_{\pm.006}$
  & $0.537_{\pm.010}$
  & $0.597_{\pm.039}$
  & $0.767_{\pm.029}$ \\
% Modality-dropout baseline (zero tab branch w.p. 0.5 during training);
% NOT SMIL -- our impl is plain dropout, not Ma et al. 2021's meta-learning.
Modality dropout (844)
  & MRI+Tab
  & $0.800_{\pm.006}$
  & $0.701_{\pm.005}$
  & $0.682_{\pm.006}$
  & $0.561_{\pm.006}$
  & $0.639_{\pm.010}$
  & $0.762_{\pm.013}$ \\
Modality dropout (844)
  & MRI+Tab+PET
  & $0.799_{\pm.020}$
  & $0.693_{\pm.017}$
  & $0.669_{\pm.008}$
  & $0.550_{\pm.020}$
  & $0.653_{\pm.055}$
  & $0.734_{\pm.020}$ \\
HeMIS~\cite{havaei2016hemis} (844)
  & MRI+Tab
  & $0.775_{\pm.008}$
  & $0.688_{\pm.032}$
  & $0.699_{\pm.031}$
  & $0.542_{\pm.053}$
  & $0.486_{\pm.069}$
  & $\mathbf{0.889_{\pm.010}}$ \\
% Graph-smoothness regularization baseline (Weston et al. 2008 semi-supervised
% embedding): k=10 NN graph on MRI features + pairwise L2 penalty on neighbour
% embeddings added to the classification loss. NOT Ou et al. / Graph-SLC.
% Per-seed AUC 0.760/0.790/0.746.
Graph-smoothness reg.~\cite{weston2008deep} (844)
  & MRI+Tab
  & $0.753_{\pm.014}$
  & $0.667_{\pm.020}$
  & $0.650_{\pm.029}$
  & $0.520_{\pm.023}$
  & $0.597_{\pm.020}$
  & $0.736_{\pm.058}$ \\
% Graph-SLC faithful (Ou et al. 2024), post tab-leak-fix; tab zeroed at inference.
Graph-SLC~\cite{ou2024graph} (844)
  & MRI+Tab
  & $0.802_{\pm.016}$
  & $0.700_{\pm.026}$
  & $0.695_{\pm.013}$
  & $0.559_{\pm.038}$
  & $0.576_{\pm.098}$
  & $0.824_{\pm.048}$ \\
\midrule
% ---- Bottom block: 378 tab-complete subjects only ----
Fusion (pairs only) (378)
  & MRI+Tab
  & $0.750_{\pm.022}$
  & $0.653_{\pm.002}$
  & $0.654_{\pm.003}$
  & $0.493_{\pm.005}$
  & $0.486_{\pm.026}$
  & $0.819_{\pm.022}$ \\
Suk et al.~\cite{suk2014hierarchical} (378)$^{\dagger}$
  & MRI+Tab
  & $0.689_{\pm.063}$
  & $0.628_{\pm.088}$
  & $0.590_{\pm.140}$
  & $0.610_{\pm.035}$
  & $0.710_{\pm.121}$
  & $0.545_{\pm.292}$ \\
\midrule
% ---- Published architectures (different backbone; MRI-only inference; reference) ----
DiaMond~\cite{li2025diamond}
  & MRI+PET
  & $0.641_{\pm.023}$ & $0.601_{\pm.019}$ & $0.602_{\pm.027}$
  & $0.415_{\pm.018}$ & $0.403_{\pm.055}$ & $0.798_{\pm.084}$ \\
Wang et al.~\cite{wang2024joint}
  & MRI
  & $0.752_{\pm.071}$ & $0.677_{\pm.067}$ & $0.670_{\pm.054}$
  & $0.521_{\pm.092}$ & $0.542_{\pm.148}$ & $0.811_{\pm.016}$ \\
HyperFusion~\cite{duenias2025hyperfusion}
  & MRI+Tab
  & $0.750_{\pm.010}$ & $0.592_{\pm.051}$ & $0.586_{\pm.064}$
  & $0.344_{\pm.134}$ & $0.292_{\pm.153}$ & $0.892_{\pm.051}$ \\
IC-MKD~\cite{kwak2025icmkd}
  & MRI+PET
  & $0.763_{\pm.004}$ & $0.648_{\pm.012}$ & $0.654_{\pm.010}$
  & $0.483_{\pm.021}$ & $0.451_{\pm.035}$ & $0.845_{\pm.011}$ \\
\bottomrule
\end{tabular}
\end{sidewaystable}

Table~\ref{tab:main_results} summarises held-out performance on the 177-subject ADNI~\cite{ADNIdataset} test set. The MRI-only baseline attains AUC $=0.789\pm0.016$.
Training a conventional fusion model on the 378-subject fully paired subset (Fusion, pairs only) does not improve AUC, indicating that restricting training to paired subjects is suboptimal.
With the same tabular pairing rate, PANDA (MRI+Tab) yields comparable AUC ($0.790\pm0.012$; $p_\mathrm{adj}=1.00$ vs.\ MRI-only) while improving balanced accuracy (0.692) and macro-F1 (0.689), with higher CN recall (0.822 vs.\ 0.749).

Adding PET prototypes (PANDA: MRI+Tab+PET, $r_\mathrm{PET}=0.19$) increases AUC to $0.851\pm0.010$ (not significant after Holm--Bonferroni, $p_\mathrm{adj}=0.119$) and improves AD recall from 0.632 to 0.750. Incorporating handwriting prototypes from the external DARWIN cohort~\cite{cilia2022diagnosing} (PANDA: MRI+Tab+PET+HW, $r_\mathrm{HW}=0$) further increases AUC to $0.868\pm0.020$ ($+7.9$\,pp vs.\ MRI-only; $p_\mathrm{adj}=0.011$) and raises CN recall to 0.863, demonstrating effective transfer of class geometry from a cohort with zero subject overlap.

Among baselines, Graph-SLC (a graph-embedded latent-space clustering method)~\cite{ou2024graph} achieves AUC $=0.802\pm0.016$, and modality-dropout variants are similar (0.800 and 0.799; $p=0.608$ vs.\ MRI-only). Although these methods slightly exceed PANDA (MRI+Tab) in AUC, they do not mitigate scanner-associated error (1.5\,T CN false-positive rate $41.4$--$47.8\%$ vs.\ $52.2\%$ for MRI-only), whereas PANDA with tabular anchoring reduces it to $27.9\%$ (Table~\ref{tab:scanner}). Knowledge distillation~\cite{hu2020knowledge}, HeMIS~\cite{havaei2016hemis}, and graph-smoothness regularisation~\cite{weston2008deep} perform at or below the MRI-only baseline. Additional published architectures evaluated under our protocol (DiaMond, a bi-modal MRI--PET vision transformer~\cite{li2025diamond}; the diagnosis network of Wang et al.~\cite{wang2024joint}; HyperFusion, a tabular-conditioned hypernetwork~\cite{duenias2025hyperfusion}; and IC-MKD, incomplete cross-modal mutual knowledge distillation~\cite{kwak2025icmkd}) also underperform PANDA in AUC.

\begin{figure}[t!]
\centering
\includegraphics[width=0.9\linewidth]{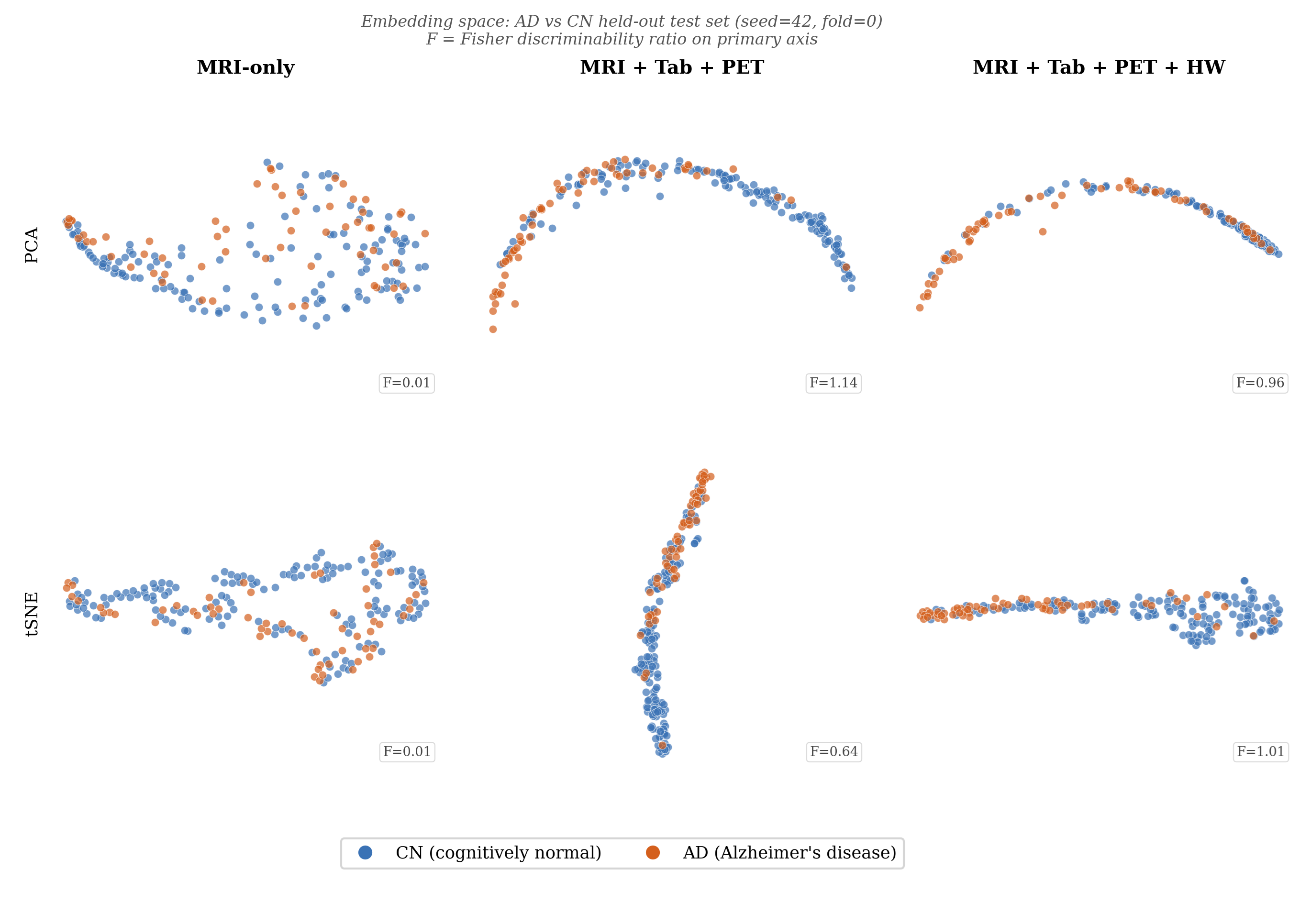}
\caption{Held-out test-set MRI embeddings (seed 42, fold 0) under PCA and t-SNE, for MRI-only, MRI+Tab+PET, and the full MRI+Tab+PET+HW model. $F$ is the Fisher discriminability ratio of CN vs.\ AD on the primary axis of each projection. Progressive auxiliary alignment reorganises the unsupervised embedding geometry into a more clearly class-separated arrangement: $F$ rises from $0.01$ for MRI-only to $1.14$ (PCA)\,/\,$0.64$ (t-SNE) for MRI+Tab+PET and $0.96$ (PCA)\,/\,$1.01$ (t-SNE) for the full MRI+Tab+PET+HW model. The projections illustrate this qualitative reorganisation; quantitative discriminability is reported by test AUC in Table~\ref{tab:main_results}.}
\label{fig:embedding_comparison}
\end{figure}

Fig.~\ref{fig:embedding_comparison} provides a qualitative visualisation of representation changes.

% ── 4.5 Scanner-stratified analysis ──────────────────────────────────────────
\subsection{Scanner-Stratified Analysis}
\label{sec:scanner}

% ---- TABLE: Scanner stratification ----
\begin{table}[t]
\centering
\footnotesize
\caption{Scanner field-strength stratification.
1.5\,T: $n = 61$ (AD $= 24$, CN $= 37$);
3\,T: $n = 116$ (AD $= 24$, CN $= 92$).
FP\,\% = CN misclassified as AD.
Mean\,$\pm$\,SD across 3 seeds.
Bold: best per column.
$^{\dagger}$Suk et al.\ is evaluated on the paired subset: 1.5\,T is still
$n = 61$, but its 3\,T group is $n = 14$ (not 116), so its 3\,T column is not
directly comparable to the other rows. The final block lists independently-published
architectures with different backbones, shown for reference and excluded from the
per-column bold.}
\label{tab:scanner}
\setlength{\tabcolsep}{4pt}
\renewcommand{\arraystretch}{1.2}
\begin{tabular}{@{}lcccc@{}}
\toprule
\textbf{Method}
  & \makecell{\textbf{1.5T AUC} $\uparrow$}
  & \makecell{\textbf{1.5T FP\,\%} $\downarrow$}
  & \makecell{\textbf{3T AUC} $\uparrow$}
  & \makecell{\textbf{3T FP\,\%} $\downarrow$} \\
\midrule
PANDA: MRI+Tab
  & $0.703_{\pm.009}$
  & $27.9_{\pm3.4}$
  & $0.804_{\pm.012}$
  & $13.8_{\pm2.2}$ \\
PANDA: MRI+Tab+PET
  & $0.736_{\pm.003}$
  & $39.6_{\pm4.5}$
  & $0.886_{\pm.020}$
  & $15.6_{\pm5.9}$ \\
\textbf{PANDA: MRI+Tab+PET+HW}
  & $\mathbf{0.783_{\pm.013}}$
  & $27.0_{\pm6.6}$
  & $\mathbf{0.894_{\pm.019}}$
  & $\mathbf{8.3_{\pm1.8}}$ \\
\addlinespace
MRI-only Baseline
  & $0.695_{\pm.009}$
  & $52.2_{\pm11.4}$
  & $0.814_{\pm.022}$
  & $14.1_{\pm2.3}$ \\
% KD (Hu et al. 2020), scanner-stratified (n=61/116).
Knowledge distillation~\cite{hu2020knowledge}
  & $0.678_{\pm.029}$
  & $41.4_{\pm1.3}$
  & $0.800_{\pm.006}$
  & $15.9_{\pm4.6}$ \\
% Modality-dropout baseline, scanner-stratified (n=61/116).
Modality dropout (Tab)
  & $0.699_{\pm.012}$
  & $41.4_{\pm2.5}$
  & $0.825_{\pm.013}$
  & $16.7_{\pm1.4}$ \\
Modality dropout (Tab+PET)
  & $0.717_{\pm.012}$
  & $47.8_{\pm8.9}$
  & $0.816_{\pm.023}$
  & $18.1_{\pm1.0}$ \\
% HeMIS (Havaei et al. 2016), scanner-stratified (n=61/116).
HeMIS~\cite{havaei2016hemis}
  & $0.665_{\pm.019}$
  & $\mathbf{17.1_{\pm4.6}}$
  & $0.823_{\pm.005}$
  & $8.7_{\pm1.8}$ \\
% Graph-smoothness regularization baseline, scanner-stratified (n=61/116).
Graph-smoothness reg.~\cite{weston2008deep}
  & $0.653_{\pm.022}$
  & $55.0_{\pm11.3}$
  & $0.762_{\pm.018}$
  & $14.9_{\pm3.6}$ \\
Graph-SLC~\cite{ou2024graph}
  & $0.703_{\pm.032}$
  & $34.2_{\pm7.1}$
  & $0.829_{\pm.029}$
  & $10.9_{\pm4.1}$ \\
\midrule
% Suk et al. (2014), paired subset: 1.5T n=61, 3T n=14 (75 total; not the full 116); see dagger note.
Suk et al.~\cite{suk2014hierarchical}$^{\dagger}$
  & $0.682_{\pm.077}$
  & $45.0_{\pm29.4}$
  & $0.707_{\pm.108}$
  & $47.6_{\pm29.4}$ \\
\midrule
% ---- Published architectures (different backbone; reference) ----
DiaMond~\cite{li2025diamond}
  & $0.494_{\pm.039}$
  & $38.7_{\pm12.5}$
  & $0.682_{\pm.025}$
  & $12.7_{\pm8.1}$ \\
Wang et al.~\cite{wang2024joint}
  & $0.667_{\pm.098}$
  & $44.1_{\pm5.6}$
  & $0.762_{\pm.082}$
  & $8.7_{\pm3.9}$ \\
HyperFusion~\cite{duenias2025hyperfusion}
  & $0.630_{\pm.011}$
  & $25.2_{\pm10.0}$
  & $0.798_{\pm.011}$
  & $5.1_{\pm3.1}$ \\
IC-MKD~\cite{kwak2025icmkd}
  & $0.672_{\pm.042}$
  & $46.8_{\pm3.4}$
  & $0.784_{\pm.005}$
  & $2.9_{\pm0.5}$ \\
\bottomrule
\end{tabular}
\end{table}

Table~\ref{tab:scanner} reports performance stratified by scanner field strength. The MRI-only baseline exhibits pronounced field-strength sensitivity, with AUC $=0.695$ at 1.5\,T versus 0.814 at 3\,T, and a 1.5\,T CN false-positive rate of 52.2\%. This behaviour is consistent with class-imbalanced acquisition (38.6\% of AD vs.\ 20.7\% of CN acquired at 1.5\,T in the training cohort), which permits a field-strength shortcut.

Tabular prototype alignment (PANDA: MRI+Tab) primarily reduces scanner-associated false positives, lowering the 1.5\,T FP\,\% from 52.2\% to 27.9\% with minimal change in AUC. Adding PET prototypes (PANDA: MRI+Tab+PET) increases discrimination, particularly at 3\,T (AUC 0.886), but yields a higher 1.5\,T FP\,\% (39.6\%), consistent with sparse PET pairing limiting its influence on the 1.5\,T subgroup.

\begin{figure}[t!]
\centering
\includegraphics[width=\linewidth]{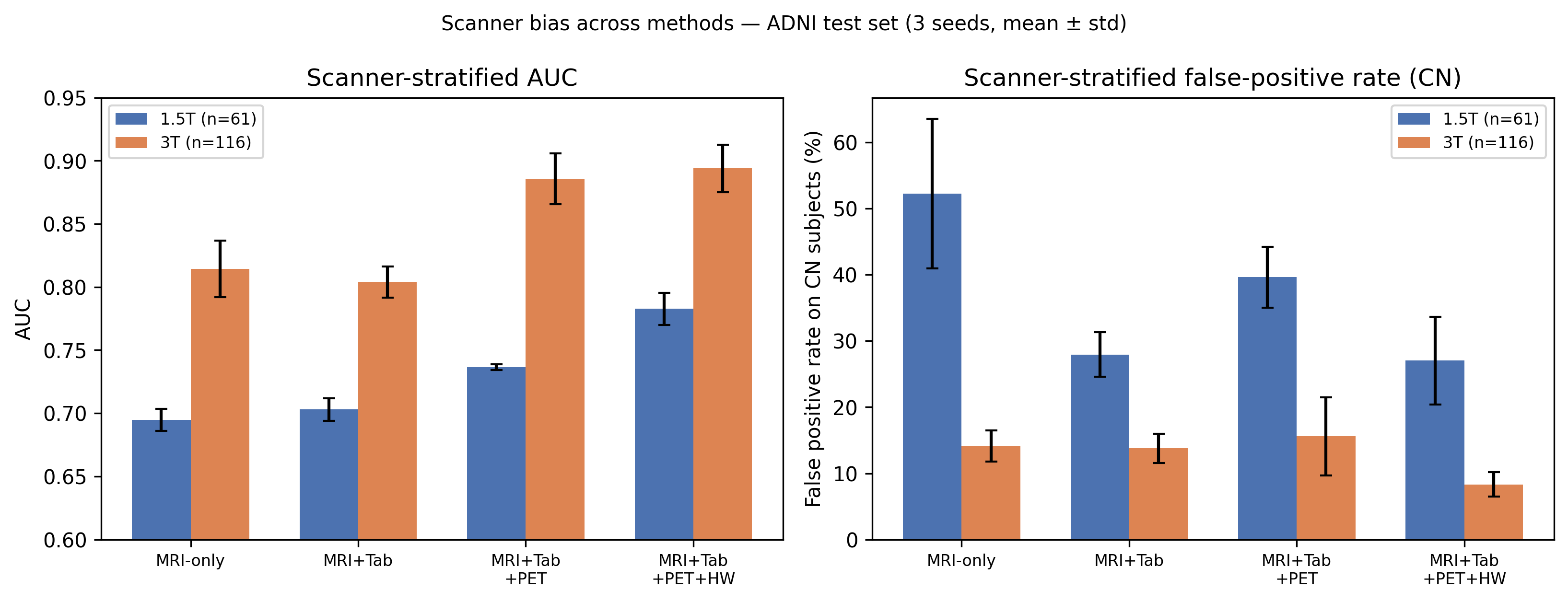}
\caption{Scanner-stratified AUC and CN false-positive rate across methods
(3 seeds, mean $\pm$ std). The full model (MRI+Tab+PET+HW) attains the highest AUC at both field strengths and the lowest 3T FP\%, reducing 1.5T FP\% to 27.0\%, without using scanner labels during training.}
\label{fig:scanner_bias}
\end{figure}

\begin{figure}[t!]
\centering
\includegraphics[width=\linewidth]{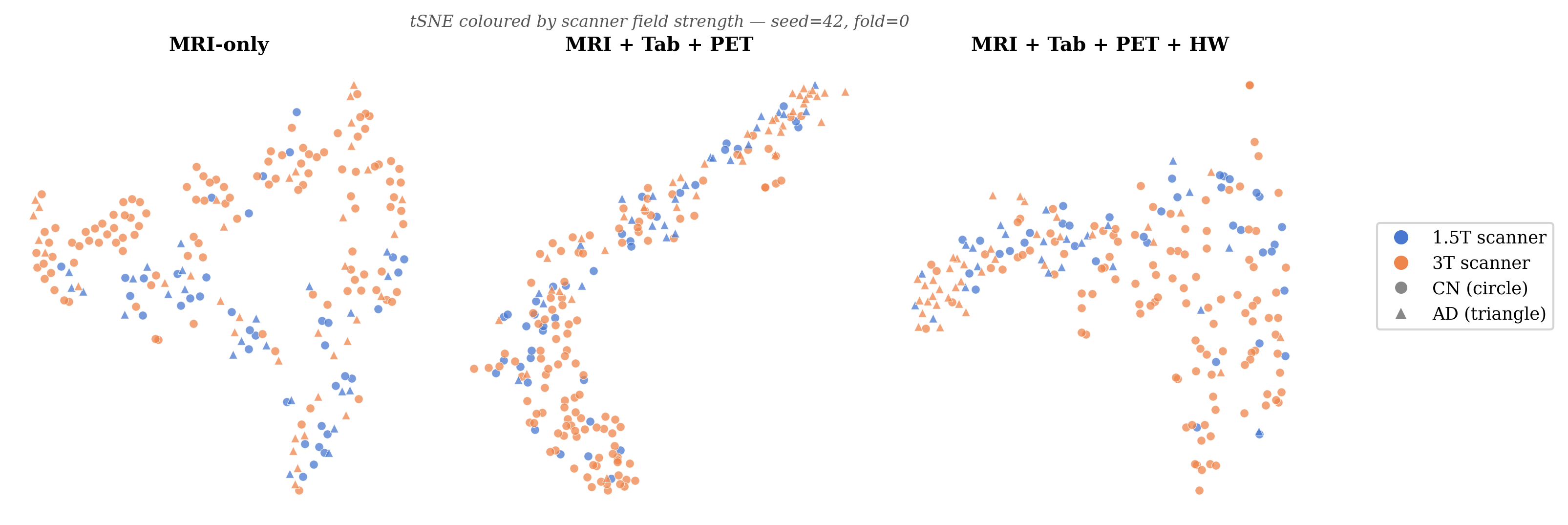}
\caption{t-SNE of held-out test-set MRI embeddings (seed 42, fold 0)
coloured by scanner field strength (1.5\,T vs.\ 3\,T) rather than by
diagnosis; marker shape indicates CN/AD. Progressive auxiliary alignment
(MRI-only $\to$ MRI+Tab+PET $\to$ MRI+Tab+PET+HW) visibly interleaves the
two scanner populations, consistent with the false-positive-rate reduction
in Fig.~\ref{fig:scanner_bias}.}
\label{fig:tsne_scanner}
\end{figure}

The full model (MRI+Tab+PET+HW) achieves the strongest overall stratified profile (1.5\,T AUC $=0.783$, FP\,\% $=27.0$; 3\,T AUC $=0.894$, FP\,\% $=8.3$). HeMIS attains a lower 1.5\,T FP\,\% (17.1\%) but does so with substantially reduced AD sensitivity (Rec-AD $=0.486$; Table~\ref{tab:main_results}). The handwriting anchor, despite having no ADNI subjects, provides complementary scanner-bias mitigation to the PET-driven AUC gains, yielding the most balanced trade-off.
Fig.~\ref{fig:scanner_bias} plots these four columns directly; Fig.~\ref{fig:tsne_scanner}
shows the corresponding t-SNE embeddings coloured by field strength rather than by diagnosis, making the 1.5\,T/3\,T mixing visible per method.

% ── 4.6 Pairing rate sensitivity ─────────────────────────────────────────────
\subsection{Ablations}
\paragraph{Pairing Rate Sensitivity}
\label{sec:pair_rate}

\begin{figure}[t!]
\centering
\includegraphics[width=0.65\linewidth]{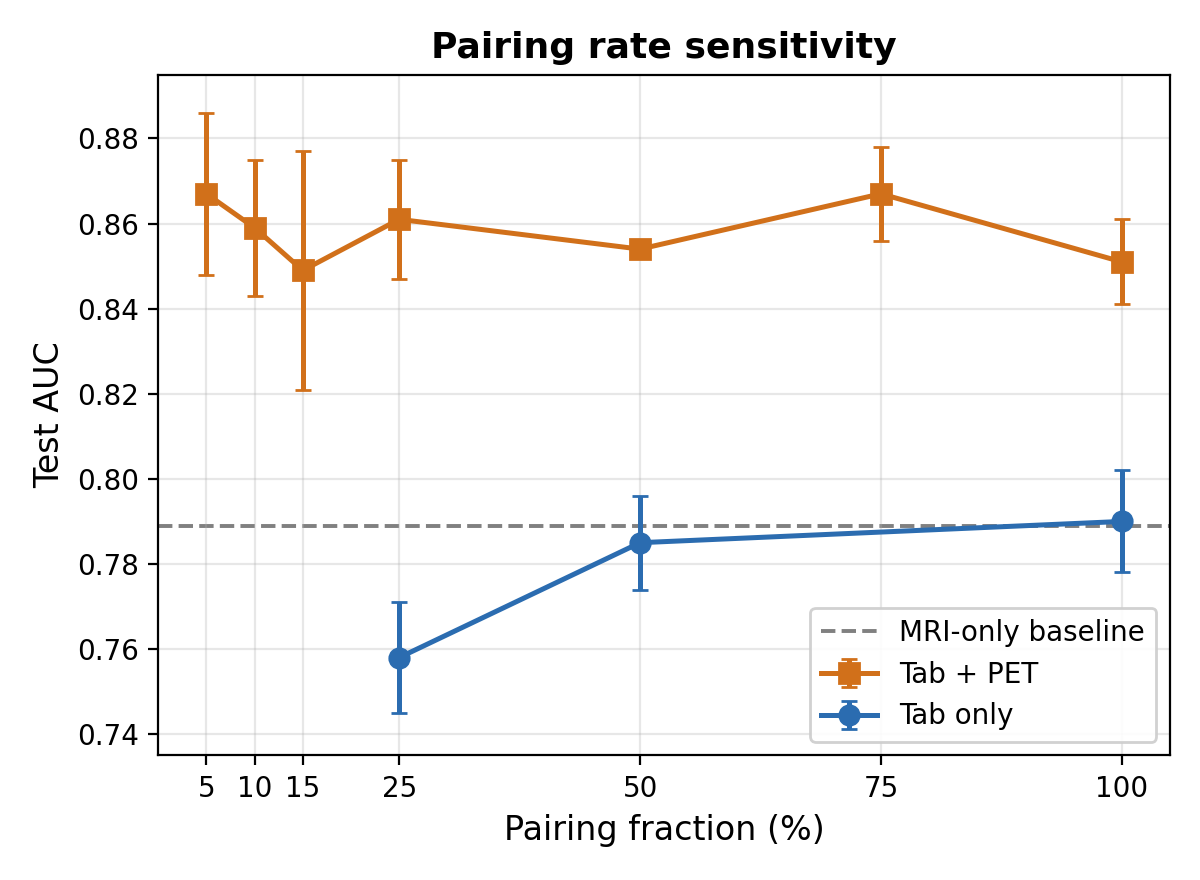}
\caption{Test AUC vs.\ pairing fraction $r$ for Tab-only and Tab+PET
alignment (mean $\pm$ std). Dashed line: MRI-only baseline ($0.789$).
Tab+PET AUC is flat within seed noise across $r$; Tab-only degrades below
$25\%$.}
\label{fig:rate_ablation}
\end{figure}

Fig.~\ref{fig:rate_ablation} reports held-out test AUC as the paired fraction $r$ is reduced from 100\% to 5\%.

% ---- figure: Pairing rate ablation ----

For tabular-only alignment, AUC is stable down to $r=50\%$ (0.790 to 0.785) but decreases at $r=25\%$ (0.758; $\approx 95$ paired subjects), which corresponds to approximately 40 AD subjects per fold for prototype estimation.
For the joint Tab+PET anchor, AUC shows no systematic dependence on pairing rate: across $r\in\{100\%,\ldots,5\%\}$ it remains within 0.849--0.867, with overlapping seed variance. Thus, reduced pairing does not improve performance, but demonstrates that full pairing is not required: $r\approx 5$--$10\%$ matches the fully paired configuration within seed noise.
This behaviour is consistent with an anchor-quality effect under finite-sample prototype estimation. For tabular prototypes, the inter-class cosine similarity (AD vs.\ CN) becomes more negative as pairing decreases (mean cosine: $-0.649$ at 100\%, $-0.731$ at 25\%, $-0.760$ at 10\%), indicating increased empirical class separation. PET prototypes remain maximally anti-parallel by construction (cosine $=-1.000$) and therefore do not contribute to this trend. A plausible explanation is that the full paired pool ($n=378$) contains a wide severity spectrum, including borderline cases that shift class means toward each other. Subsampling can preferentially yield more homogeneous within-class subsets, increasing empirical prototype separation and providing a more stable alignment target during Stage~2.

% ── Optimal-transport pseudo-pairing ablation ────────────────────────────────
\paragraph{Optimal-Transport Pseudo-Pairing}
Stage~1 contrastive alignment uses only genuinely paired subjects (tabular: 378/844; PET: $\approx$158/844), while handwriting contributes external class prototypes without per-subject pairing. To test whether unpaired MRI subjects can benefit from an additional alignment signal, we introduce an intermediate step that periodically (every 5 epochs) computes a Sinkhorn optimal-transport coupling between the MRI embedding distribution (all 844 subjects) and each auxiliary embedding distribution using cost $1-\cos(\cdot,\cdot)$. The resulting soft couplings are used as additional (pseudo-paired) InfoNCE supervision, thereby achieving an AUC score of $0.862\pm0.014$, which is statistically indistinguishable from the baseline $0.868\pm0.020$ and within our equivalence band. 

We therefore treat optimal-transport pseudo-pairing as a neutral ablation under the pairing densities considered, indicating that class-level prototype anchoring captures most of the transferable cross-modal geometry in this setting.

% ── Backbone generalizability ────────────────────────────────────────────────
\subsection{Encoder Trainability and Backbone Generalizability}
\label{sec:backbone}
Our primary MRI backbone is a MedicalNet ResNet-18 in which the first two residual stages are frozen at segmentation-pretrained weights. As a stronger fully trainable alternative, we consider the Conv5-FC3 encoder of Wen et al.~\cite{wen2020convolutional} (autoencoder warm start on the same ADNI pool; end-to-end training). On the same $n=177$ evaluation protocol, Conv5-FC3 improves the MRI-only baseline AUC from $0.789\pm0.016$ to $0.881\pm0.009$. The largest difference occurs at 1.5\,T (AUC $0.695$ vs.\ 0.800; CN FP\,\% $52.3$ vs.\ 33.3), consistent with limited trainability increasing reliance on scanner-correlated features.
As presented in Table~\ref{tab:backbone}, adding tabular alignment leaves overall AUC essentially unchanged (0.881 to 0.883) but improves scanner robustness, reducing CN FP\,\% at 1.5\,T from 33.3 to 19.8 and at 3\,T from 8.7 to 4.3. Incorporating the full four-way anchor further improves discrimination (AUC $=0.893\pm0.003$) and yields the lowest false-positive rates among the evaluated configurations.

These results indicate that prototype alignment transfers across encoder architectures and that additional auxiliary anchors can yield complementary gains. %Unless otherwise stated, all other experiments use the MedicalNet backbone.

\begin{table}[t]
\centering
\footnotesize
\caption{Backbone generalizability ($n = 177$). Prototype alignment re-run with
the fully-trainable Conv5-FC3 encoder~\cite{wen2020convolutional}.
Mean\,$\pm$\,SD across 3 seeds; bold: best per column.
$^{\dagger}$MedicalNet four-way with second stage checkpoints selected based on lowest validation loss with highest validation AUC.}
\label{tab:backbone}
\setlength{\tabcolsep}{4pt}
\renewcommand{\arraystretch}{1.2}
\begin{tabular}{@{}>{\raggedright\arraybackslash}p{0.36\linewidth}ccccc@{}}
\toprule
\textbf{Model}
  & \makecell{\textbf{AUC}}
  & \makecell{\textbf{1.5T AUC}}
  & \makecell{\textbf{1.5T FP\,\%}}
  & \makecell{\textbf{3T AUC}}
  & \makecell{\textbf{3T FP\,\%}} \\
\midrule
Conv5-FC3 (MRI-only)~\cite{wen2020convolutional}
  & $0.881_{\pm.009}$ & $0.800_{\pm.007}$ & $33.3_{\pm7.7}$
  & $0.911_{\pm.010}$ & $8.7_{\pm1.5}$ \\
PANDA\,+\,Conv5-FC3 (MRI+Tab)
  & $0.883_{\pm.004}$ & $0.799_{\pm.011}$ & $19.8_{\pm9.2}$
  & $0.912_{\pm.008}$ & $4.3_{\pm0.9}$ \\
PANDA, MedicalNet (MRI+Tab+PET+HW)$^{\dagger}$
  & $0.875_{\pm.015}$ & $0.789_{\pm.017}$ & $24.3_{\pm2.2}$
  & $0.899_{\pm.012}$ & $7.2_{\pm2.0}$ \\
\textbf{PANDA\,+\,Conv5-FC3 (MRI+Tab+PET+HW)}$^{\dagger}$
  & $\mathbf{0.893_{\pm.003}}$ & $\mathbf{0.816_{\pm.010}}$ & $\mathbf{18.0_{\pm7.1}}$
  & $\mathbf{0.916_{\pm.006}}$ & $\mathbf{4.0_{\pm1.4}}$ \\
\bottomrule
\end{tabular}
\end{table}

% ── 4.7 MCI spectrum inference ───────────────────────────────────────────────
\subsection{Zero-Shot MCI Severity Inference}
\label{sec:mci}
\begin{figure}[t!]
\centering
\includegraphics[width=\linewidth]{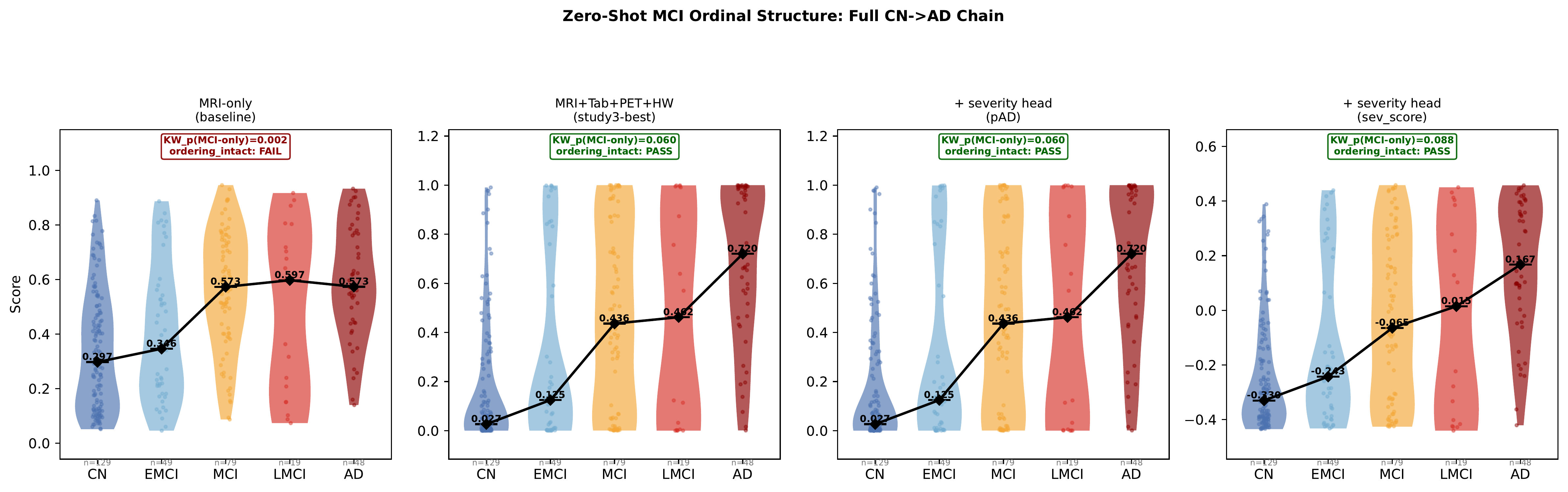}
\caption{Zero-shot MCI ordinal structure across the full CN$\to$AD chain
(not just EMCI/MCI/LMCI), for the MRI-only baseline, MRI+Tab+PET+HW
(the full model), and the severity-head extension on both $p_\mathrm{AD}$ and
\texttt{sev\_score}. The connecting line through group medians makes \texttt{ordering\_intact}
visible directly: the MRI-only panel does not rise at the AD step (AD median
$0.573$ ties the MCI median), failing the monotone chain, whereas the joint
MRI+Tab+PET+HW model and both severity-head panels rise monotonically from CN
to AD.}
\label{fig:mci_violin}
\end{figure}

The held-out MCI cohort comprises 147 subjects (EMCI: 49, MCI: 79, LMCI: 19). For anchored models, applying the prototype-distance rule to each subject's MRI projection yields a continuous $p_\mathrm{AD}^{(i)}$; for the severity-head extension (Section~\ref{sec:severity_head}), we additionally compute $\mathrm{sev\_score}^{(i)}$ via $h_\mathrm{sev}$. We report two complementary statistic families (Table~\ref{tab:mci}). Primary analyses include a 3-group Kruskal--Wallis test over EMCI/MCI/LMCI and one-sided Mann--Whitney tests on adjacent stage pairs. We also report \texttt{ordering\_intact}, a monotone-median check over CN\,$<$\,EMCI\,$<$\,MCI\,$<$\,LMCI\,$<$\,AD (computed with CN and AD included), which evaluates whether MCI falls at the expected ordinal position rather than only whether sub-stages differ. Fig.~\ref{fig:mci_violin} shows the full distributions.

\begin{table}[t]
\centering
\caption{Zero-shot MCI ordinal analysis under the primary-analysis hierarchy
(primary: 3-group KW restricted to EMCI/MCI/LMCI, and one-sided adjacent-pair
Mann--Whitney; \texttt{ordering\_intact}: full monotone chain
CN$<$EMCI$<$MCI$<$LMCI$<$AD on medians, computed over all 177 test subjects
plus the 147 MCI subjects). ns: $p \geq 0.05$.}
\label{tab:mci}
\setlength{\tabcolsep}{3pt}
\renewcommand{\arraystretch}{1.2}
\begin{tabular}{@{}lccccc@{}}
\toprule
\textbf{Method} & \textbf{Score}
  & \makecell{KW $p$\\(MCI-only)}
  & \makecell{MW $p$\\EMCI$<$MCI}
  & \makecell{MW $p$\\MCI$<$LMCI}
  & \textbf{ordering\_intact} \\
\midrule
MRI-only baseline & $p_\mathrm{AD}$
  & $0.0023$ & $<0.001$ & $0.788$ (ns) & \textbf{FAIL} \\
PANDA: MRI+Tab+PET+HW & $p_\mathrm{AD}$
  & $0.060$ (ns) & $0.007$ & $0.677$ (ns) & \textbf{PASS} \\
PANDA + severity head & $p_\mathrm{AD}$
  & $0.060$ (ns) & $0.007$ & $0.677$ (ns) & \textbf{PASS} \\
PANDA + severity head & sev\_score
  & $0.088$ (ns) & $0.010$ & $0.637$ (ns) & \textbf{PASS} \\
\bottomrule
\end{tabular}
\end{table}

The MRI-only baseline separates EMCI/MCI/LMCI on the primary 3-group test (Kruskal--Wallis $p=0.0023$) but fails \texttt{ordering\_intact}: the AD median ties the MCI median ($0.573$), yielding a non-monotone chain (CN/EMCI/MCI/LMCI medians $0.297<0.346<0.573<0.597$, with AD at $0.573$). In contrast, the four-way joint model (MRI+Tab+PET+HW; Table~\ref{tab:main_results}) passes \texttt{ordering\_intact} with monotone medians $0.027<0.125<0.436<0.462<0.720$ across CN/EMCI/MCI/LMCI/AD, despite only weak evidence for MCI sub-stage separation (KW $p=0.060$).

The severity-head extension (Section~\ref{sec:severity_head}) adds an explicit severity readout while leaving test AUC unchanged ($0.868\pm0.020$; Section~\ref{sec:main_results}). On this cohort, \texttt{ordering\_intact} passes for both $p_\mathrm{AD}$ (KW $p=0.060$) and \texttt{sev\_score} (KW $p=0.088$). Neither KW test reaches conventional significance; accordingly, we do not claim sharp separation of MCI sub-stages. The supported conclusion is that the CN$\to$AD median ordering is monotone for the joint and severity-head models, whereas it is not for the MRI-only baseline. \texttt{sev\_score} correlates strongly with $p_\mathrm{AD}$ (Pearson $r=0.941$) but is not a trivial reparameterization ($r\neq1$; Section~\ref{sec:discussion}).

% ── 4.8 Cross-domain generalisation: TCGA ────────────────────────────────────
We next evaluate the generality of prototype alignment under a shift in both data domain and objective, moving from ADNI diagnosis to TCGA-Lung survival prediction.
\subsection{Second Application: Cross-Domain Generalisation to TCGA-Lung Survival}
\label{sec:tcga}

% ---- TABLE: TCGA ----
\begin{table}[t]
\centering
\caption{TCGA-Lung generalisation. Binary 2yr OS: test $n = 119$.
Cox PH: test $n = 171$.
Mean\,$\pm$\,SD across 3 seeds.
Bold: best per column.
$^\dagger$ RNA used at inference.}
\label{tab:tcga}
\setlength{\tabcolsep}{3pt}
\renewcommand{\arraystretch}{1.2}
\begin{tabular*}{\linewidth}{@{\extracolsep{\fill}}lccc}
\toprule
\textbf{Method}
  & \makecell{\textbf{Binary AUC}}
  & \makecell{\textbf{Binary F1}}
  & \makecell{\textbf{Cox C-index}} \\
\midrule
WSI-only
  & $0.593_{\pm.020}$
  & $0.389_{\pm.021}$
  & $0.460_{\pm.003}$ \\
Full Fusion (WSI+RNA)$^\dagger$
  & $0.591_{\pm.008}$
  & $0.348_{\pm.032}$
  & $0.453_{\pm.002}$ \\
Paired-only 50\% (no proto.\ transfer)
  & $0.579_{\pm.030}$
  & $0.338_{\pm.023}$
  & $0.527_{\pm.029}$ \\
Paired-only 25\% (no proto.\ transfer)
  & $0.576_{\pm.038}$
  & $0.376_{\pm.056}$
  & $0.513_{\pm.023}$ \\
\midrule
PANDA 25\% RNA
  & $0.593_{\pm.026}$
  & $0.336_{\pm.078}$
  & $0.529_{\pm.018}$ \\
PANDA 50\% RNA
  & $0.612_{\pm.018}$
  & $0.380_{\pm.053}$
  & $0.519_{\pm.038}$ \\
\textbf{PANDA 100\% RNA}
  & $\mathbf{0.628_{\pm.007}}$
  & $\mathbf{0.495_{\pm.015}}$
  & $\mathbf{0.550_{\pm.016}}$ \\
\bottomrule
\end{tabular*}
\end{table}

\begin{figure}[t!]
\centering
\includegraphics[width=\linewidth]{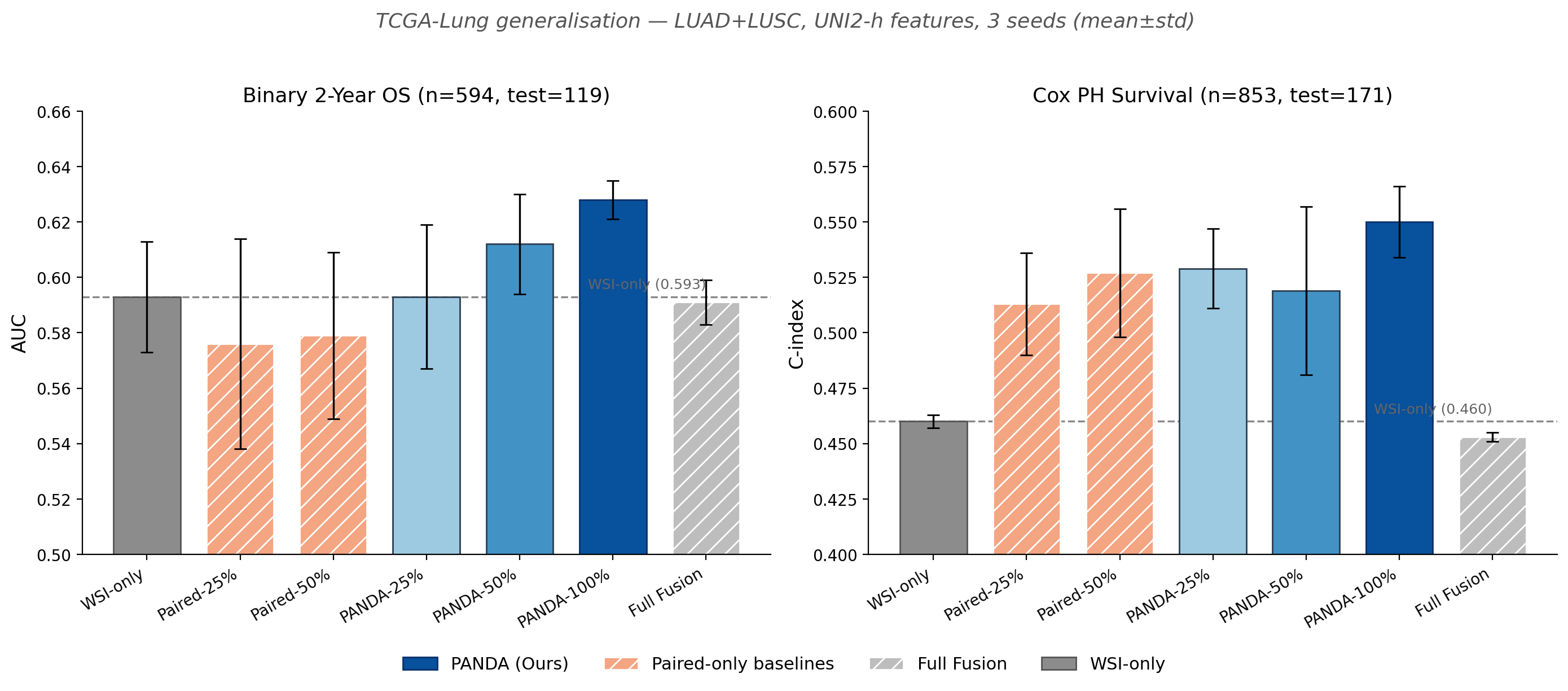}
\caption{TCGA-Lung generalisation (3 seeds, mean $\pm$ std). \textbf{Left:}
binary 2-year overall survival AUC. \textbf{Right:} Cox proportional-hazards
C-index. PANDA (prototype alignment) at 100\% RNA pairing outperforms
both WSI-only and Full Fusion on both tasks without using RNA at inference;
Paired-only baselines at matched RNA fractions fall below PANDA at the
same fraction, isolating the gain to the prototype-transfer mechanism
rather than RNA availability alone.}
\label{fig:tcga}
\end{figure}

Table~\ref{tab:tcga} and Fig.~\ref{fig:tcga} summarise TCGA results. The WSI-only baseline attains AUC $=0.593\pm0.020$ (95\,\%\,CI [0.492, 0.689]) for binary 2-year OS and C-index $=0.460\pm0.003$ (95\,\%\,CI [0.383, 0.537]) for Cox PH, i.e., near-chance survival discrimination from H\&E morphology alone. Full Fusion (joint WSI\,+\,RNA ABMIL; RNA used at inference) does not improve over WSI-only on either endpoint (AUC $=0.591$, C-index $=0.453$), indicating unstable optimisation when trained directly on the partially paired pool ($\approx470$ subjects).

Restricting Stage~2 to the paired subset (Paired-only baselines) likewise matches or underperforms WSI-only on binary AUC (0.576--0.579), indicating that discarding unpaired subjects is detrimental. In contrast, prototype-based contrastive transfer using 100\% RNA pairing improves both endpoints without RNA at inference (AUC $=0.628\pm0.007$, 95\,\%\,CI [0.521, 0.733]; C-index $=0.550\pm0.016$, 95\,\%\,CI [0.482, 0.616]), corresponding to absolute gains of $+3.5$\,pp AUC and $+9.0$ C-index points vs.\ WSI-only. Bootstrap tests vs.\ WSI-only yield $p=0.146$ (AUC) and $p=0.059$ (C-index); with $n_\mathrm{test}=119$ (binary) and 171 (Cox), CIs remain wide and the improvements do not reach $\alpha=0.05$, although the direction is consistent across seeds and tasks.

Matched-fraction Paired-only comparisons support a prototype-transfer effect beyond pairing rate: at 25\% RNA, PANDA achieves C-index $=0.529$ vs.\ 0.513 for Paired-only 25\% ($+1.6$ pts) with identical RNA training data.

% ============================================================
% §5  DISCUSSION
% ============================================================

\section{Discussion}
\label{sec:discussion}

Across ADNI and TCGA-Lung, prototype alignment consistently converts partially paired auxiliary modalities into improvements in the primary-modality model (better calibration and scanner robustness) without requiring auxiliary inputs at inference. The effect is mediated by geometry: each auxiliary anchor reshapes the MRI/WSI embedding space, and the three anchors studied target complementary geometric failure modes.

Tabular alignment provides the most direct example of geometric regularisation. It leaves AUC approximately unchanged but improves CN recall and macro-F1 (Table~\ref{tab:main_results}), consistent with reduced within-class scatter around clinically defined cluster centres rather than added boundary signal. This restructuring also explains the large 1.5\,T false-positive reduction (Table~\ref{tab:scanner}): borderline 1.5\,T controls are pulled toward a clinically grounded CN anchor instead of remaining near the decision boundary. PET exhibits the complementary behaviour: it yields the largest single AUC gain but does not improve 1.5\,T robustness (Tables~\ref{tab:main_results}, \ref{tab:scanner}). A likely explanation is limited coverage: at 19\% pairing, the PET prototype is estimated from $\approx$30 paired AD subjects per fold and is dominated by high-quality 3\,T scans from later ADNI phases, providing limited constraint on the borderline 1.5\,T controls that drive false positives. Increasing PET pairing density, or adding an explicit 1.5\,T-robust regulariser, remains future work.

This modality-specific division of labour is also reflected in baseline behaviour. Modality-dropout baselines benefit little from PET (Table~\ref{tab:main_results}: MRI+Tab 0.800 vs.\ MRI+Tab+PET 0.799), whereas prototype alignment improves substantially (0.790$\to$0.851). We do not isolate the mechanism experimentally, but hypothesise that sparse-modality utilisation differs: concatenate-and-mask training exposes the fusion head to real PET only on the paired fraction (19\% pairing, further reduced by dropout) and cannot propagate PET-derived structure to the unpaired majority, whereas prototype alignment estimates class-level PET geometry from paired subjects and propagates it to all MRI embeddings via the alignment loss. In addition, the dropout head is trained on mixed present/zero-filled auxiliary patterns that differ from deployment, while PANDA’s classifier consumes MRI features consistently at train and test. We will test these hypotheses in future work by stratifying performance on PET-paired versus PET-unpaired test subjects.

The handwriting anchor isolates the minimal requirement for cross-cohort transfer. With $r_{\mathrm{HW}}=0$, no ADNI subject contributes to the DARWIN prototype, yet handwriting yields the strongest scanner-stratified performance (Table~\ref{tab:scanner}). This indicates that transfer does not require semantic modality alignment; it requires prototype \emph{discriminability}. A well-separated class geometry in one domain (handwriting kinematics) can constrain a structurally unrelated domain (3D MRI) when labels are shared.

The pairing-rate ablation (Figure~\ref{fig:rate_ablation}) further supports deployability: full pairing is unnecessary. The Tab+PET anchor remains within seed noise from 75\% to 5\% pairing, whereas tabular-only alignment degrades only when the paired pool falls below $\approx$95 subjects per fold, where class-mean estimates become unstable. We attribute the Tab+PET stability to prototype quality: the fully paired pool spans a broad severity spectrum, including borderline subjects that dilute class means; random subsampling more often excludes these ambiguous cases, yielding tighter and more separated prototypes (mirrored by increasing AD--CN prototype separation as pairing decreases). Under resource constraints, sampling a modest number of pairs per modality can be preferable to maximising pairing completeness.

The TCGA-Lung results are consistent with the same decoupling of alignment and classification: effects are directionally consistent across seeds and tasks but underpowered at the available test sizes (Table~\ref{tab:tcga}). Binary 2-year survival is difficult to power because censoring and thresholding discard time-to-event information, whereas the Cox C-index (risk ranking) shows a numerically larger gain ($+9.0$ points; $0.460\to0.550$) that did not reach the conventional $\alpha=0.05$ threshold ($p=0.059$; Table~\ref{tab:tcga}); the per-arm $95\%$ CIs overlap (WSI-only $[0.383,0.537]$ vs.\ PANDA $[0.482,0.616]$), so we report it as a consistent directional trend rather than a clinically established effect. Full Fusion falling to or below WSI-only on Cox is consistent with early-fusion failure under small paired pools~\cite{baltrusaitis2018multimodal}: the $\approx$470 paired subjects are insufficient to learn a reliable RNA-to-survival mapping while maintaining WSI attention quality. PANDA avoids this by separating alignment (Stage~1, paired subjects) from prediction (Stage~2, all subjects), ensuring the survival head is trained on the full WSI set.

Finally, the aligned representation encodes structure beyond binary objectives. In zero-shot MCI inference (Section~\ref{sec:mci}), unimodal CN/AD separation does not reliably embed MCI along the severity axis (the ordinal ordering is not preserved), whereas prototype alignment stabilises the CN$\to$AD ordering, and an explicit severity head—trained without any MCI supervision—recovers the same monotone ordering without AUC cost. These results indicate that severity information is already present in the MRI representation and primarily requires an appropriate readout objective.

Not all approaches to exploit this structure were supported. A geometry-transfer variant (matching the encoder’s inter-prototype angle to auxiliary pre-computed angles) achieved the highest raw metric ($0.877\pm0.018$), but a four-arm ablation (real, identity-shuffled, label-shuffled correspondence, and a gradient-severed no-op) could not distinguish it from the no-op control \emph{on downstream AUC}. The null result is confined to classification accuracy: the geometry pathway itself behaved as designed (values are mean${}\pm{}$s.d.\ across three seeds). In the real arm the inter-prototype angle converged toward the auxiliary target, with cosine moving from $-0.597\pm0.020$ to $-0.769\pm0.019$ and the geometry loss falling ${\approx}4.7\times$ ($0.412\pm0.035\to0.089\pm0.012$). In the gradient-severed no-op neither moved toward the target: the cosine stayed positive and drifted away from the (negative) auxiliary targets ($+0.227\pm0.092\to+0.370\pm0.092$), and the geometry loss did not decrease. Reshaping this angle therefore did not, on its own, translate into an AUC gain over the no-op---not that the loss had no effect on the representation. A plausible explanation is a training-exposure artefact of the dual-loader control: at every step both arms draw an additional, larger batch through the shared MRI encoder and projection head (matching compute and BatchNorm running statistics), and only the geometry-loss coefficient is zeroed in the no-op; the shared extra exposure through BatchNorm statistics can account for most of any regularisation benefit, making AUC an insensitive readout for whether the geometry loss itself matters. We did not log per-pathway gradient norms to confirm this directly, so we report the exposure-artefact account as an inference from the control's design rather than a measured quantity (Section~\ref{sec:limitations}).

\paragraph{Limitations and future work.}
\label{sec:limitations}
The evaluation is confined to ADNI for AD/CN classification; external MRI cohorts (e.g., OASIS-3~\cite{lamontagne2019oasis3}, UK~Biobank~\cite{sudlow2015ukbiobank}) are required to establish generalisability across sites and scanners. The persistent 1.5\,T/3\,T gap (27\,pp FP\,\% vs.\ 8\,pp at 3\,T, even for the best model) indicates that prototype alignment mitigates but does not eliminate scanner-related bias; deployment in 1.5\,T-dominant settings will likely require dedicated harmonisation. One contributing factor is the restricted trainability of the encoder: freezing the early backbone layers limits adaptability, while making the entire encoder trainable boosts the unimodal baseline and yields the strongest aligned performance (Section~\ref{sec:backbone}). However, backbone design itself was not systematically optimized in the main study. PET pairing at 19\% is also too sparse to enforce scanner-robust alignment at 1.5\,T; denser PET pairing (or amyloid PET) is a natural extension.

The severity-head extension (Section~\ref{sec:severity_head}) is trained on $s_\mathrm{orth}$, available for only 265/844 training subjects, and this subset is scanner-skewed (208 at $\leq2.0$\,T vs.\ 57 at $>2.0$\,T). Denser and more scanner-balanced severity coverage would strengthen conclusions beyond the current single-composite result. A within-class pairwise rank loss, explored as an alternative approach to ordinal structure, did not improve sub-stage separation and degraded AD-adjacent ordering (MCI$\to$LMCI) in 3-seed evaluation; we therefore do not pursue it further. This loss is a RankNet-style pairwise logistic term (binary cross-entropy on the predicted severity gap $\hat{s}_i-\hat{s}_j$ for each non-tied within-class pair, with no margin parameter), applied only to the AD sugnnore tbset ($n=120$; the CN population showed no usable within-class severity spread), with weight $\lambda_\mathrm{wc}=0.3$ on $s_\mathrm{orth}$ and pairs drawn from a dedicated severity loader. Compared against the severity-head baseline ($\lambda_\mathrm{wc}=0$) using the primary adjacent-pair tests, it flipped the \texttt{ordering\_intact} check from pass to fail on the severity readout, with the LMCI median falling below the MCI median. The geometry-transfer variant (Section~\ref{sec:discussion}) was indistinguishable from a no-op control \emph{in downstream AUC}---despite measurably reshaping the encoder's inter-prototype geometry---and we did not log per-pathway gradient norms to establish whether the dual-loader exposure artefact or the loss itself explains this; we therefore treat it as unconfirmed and flag this outcome for others using similar dual-loader training. Finally, loss weights ($\lambda_\mathrm{tab}$, $\lambda_\mathrm{PET}$, $\lambda_\mathrm{HW}$) were tuned at 100\% pairing, and a systematic sweep is deferred to revision.

Three directions follow directly from these findings.
\emph{Multi-cohort generalisation.} The core claim—relaxing full pairing to exploit larger training cohorts improves the primary-modality classifier—is demonstrated on ADNI and TCGA-Lung. The next step is evaluation across additional external AD cohorts (OASIS-3~\cite{lamontagne2019oasis3}, AIBL~\cite{ellis2009aibl}, UK~Biobank~\cite{sudlow2015ukbiobank}) and other partially paired benchmarks, to characterise when the approach succeeds as a general property rather than a dataset-specific effect.
\emph{Non-adversarial scanner harmonisation.} The residual 1.5\,T/3\,T gap (Section~\ref{sec:limitations}) and the failure of distribution-matching baselines (adversarial scanner discrimination and sliced-Wasserstein alignment on PET embeddings~\cite{lee2019sliced,ganin2016dann}) motivate prototype-space alternatives. Aligning per-scanner class centroids avoids fitting a discriminator and requires only class and domain labels, not dense per-domain distribution estimates. A power analysis linking harmonisation performance to the number of paired scanner samples would quantify the additional data requirements and inform future multi-site collection.
\emph{Richer anchor geometry and severity supervision.} The severity head currently relies on a single label-orthogonal composite available for 265 CN/AD subjects; denser, scanner-balanced graded scores, and anchors that represent within-class heterogeneity (mixture/distributional prototypes rather than a single class mean), are promising extensions given substantial within-AD severity variation observed in our data.

% §6 Conclusion
%\subsection{Future Work}
%\label{sec:future}

\section{Conclusion}
\label{sec:conclusion}
We introduced PANDA, a two-stage prototype-anchored alignment framework for multimodal AD classification with partial pairing. A single training procedure accommodates heterogeneous pairing densities (44.8\% tabular, 18.7\% PET, and 0\% external handwriting) while requiring only MRI at inference. On ADNI, the full model achieves AUC $=0.868$ ($+7.9$\,pp vs.\ MRI-only) and reduces the 1.5\,T CN false-positive rate by $>24$\,pp. 
The reported gains are relative to same-backbone MRI-only baselines. Conversely, a fully trainable Conv5-FC3 encoder attains AUC $=0.881$ without alignment, and PANDA improves it further to AUC $=0.893$ while approximately halving the 1.5\,T false-positive rate. Thus, across backbones, prototype alignment improves the corresponding baseline, with larger AUC gains for weaker encoders and larger robustness gains once AUC saturates.
A pairing-rate ablation shows that full pairing is not required for the joint anchor: $5$--$10\%$ pairing matches the fully paired configuration within seed noise, implying that collecting additional modalities at low overlap can be preferable to assembling a small fully paired cohort. Cross-domain evaluation on TCGA-Lung indicates that the benefits are not neuroimaging-specific.

Zero-shot analysis on held-out MCI subjects further characterises the representation: the unimodal MRI baseline does not preserve the CN$\to$AD ordinal chain, whereas the prototype-aligned model maintains monotone group medians. A linear severity head, trained on graded CN/AD scores without any MCI gradients, exposes this axis explicitly with no AUC cost, indicating that severity information is present in the aligned embedding and can be read out without MCI-specific supervision. In contrast, a class-prototype \emph{geometry-transfer} variant was not supported: despite the highest raw metric, a four-arm ablation (real, identity-shuffled, label-shuffled correspondence, and a gradient-severed no-op with matched training exposure) could not distinguish it from the no-op control, consistent with training-exposure confounding rather than a geometry-transfer effect. Overall, these results support partially paired prototype alignment as a deployment-oriented strategy for leveraging incomplete auxiliary data to improve accuracy and scanner robustness without auxiliary inputs at test time.

\section*{Code Availability}
The full training, evaluation, and analysis code for PANDA will be made publicly available upon acceptance. The datasets used are accessible through their respective repositories under their own data-use agreements: ADNI (\url{https://adni.loni.usc.edu}), TCGA via the GDC Data Portal (\url{https://portal.gdc.cancer.gov}), and the DARWIN handwriting dataset~\cite{cilia2022diagnosing}.

\section*{Acknowledgements}
The authors gratefully acknowledge the scientific support and HPC resources provided by the Erlangen National High Performance Computing Center (NHR@FAU) of the Friedrich-Alexander-Universit\"at Erlangen-N\"urnberg (FAU). The hardware is partially funded by the German Research Foundation (DFG).

Data used in preparation of this article were obtained from the Alzheimer's Disease Neuroimaging Initiative (ADNI) database (\url{https://adni.loni.usc.edu}). As such, the investigators within the ADNI contributed to the design and implementation of ADNI and/or provided data but did not participate in analysis or writing of this report. A complete listing of ADNI investigators can be found at \url{https://adni.loni.usc.edu/wp-content/uploads/how_to_apply/ADNI_Acknowledgement_List.pdf}.

\appendix
\section{Baseline Configurations}
\label{app:baselines}

\begin{table}[h]
\centering
\small
\setlength{\tabcolsep}{4pt}
\caption{Baselines and their configuration. ``Aux (train)'': auxiliary
modalities used during training only; ``Infer'': inputs required at test time.
Unless stated, ADNI baselines train on all 844 subjects and are evaluated on the
full test set ($n=177$).}
\label{tab:baselines}
\begin{tabular}{@{}>{\raggedright\arraybackslash}p{2.4cm}ccp{5.0cm}@{}}
\toprule
Method & Aux (train) & Infer & Key configuration / adaptation \\
\midrule
\multicolumn{4}{@{}l}{\emph{ADNI (AD/CN classification)}}\\
MRI-only                                  & ---           & MRI     & MedicalNet ResNet-18, cross-entropy only \\
MRI+Tab (pairs)                               & Tab           & MRI+Tab & Feature concatenation; trained on the 378 paired subjects only \\
Suk (SAE fusion)~\cite{suk2014hierarchical}    & Tab           & MRI+Tab & Stacked-autoencoder late fusion; both modalities at test; paired test ($n{=}75$) \\
Knowledge distill.~\cite{hu2020knowledge}        & Tab+PET       & MRI     & MRI student matches MRI+Tab+PET prototype teacher (KL, $T{=}4$) \\
Modality dropout                            & Tab / Tab+PET & MRI     & Random branch dropout ($p{=}0.5$); two configurations \\
HeMIS~\cite{havaei2016hemis}        & Tab+PET       & MRI     & Hetero-modal mean/variance pooling over present modalities \\
Graph-smoothness~\cite{weston2008deep}         & ---           & MRI     & Manifold regularisation; $k{=}10$ NN graph, pairwise $L_2$ penalty \\
Graph-SLC~\cite{ou2024graph}            & Tab           & MRI     & Graph-embedded latent; decoders reconstruct missing; tabular zeroed at test \\
DiaMond~\cite{li2025diamond}          & PET           & MRI     & Bi-modal MRI+PET ViT (bi-attention), run MRI-only \\
Wang et al.~\cite{wang2024joint}          & PET           & MRI     & Diagnosis network; PET-synthesis branch omitted \\
HyperFusion~\cite{duenias2025hyperfusion} & Tab           & MRI     & Hypernetwork conditions MRI on tabular (age, sex, APOE) \\
IC-MKD~\cite{kwak2025icmkd}          & PET           & MRI     & MRI+PET teacher $\to$ MRI student; reimplemented (no official code) \\
CCSW~\cite{lee2019sliced}          & Tab           & MRI     & Contrastive class-conditional sliding-window alignment \\
\midrule
\multicolumn{4}{@{}l}{\emph{TCGA-Lung (survival)}}\\
Full Fusion                                  & RNA           & WSI+RNA & Joint WSI+RNA ABMIL with RNA at inference (complete-data upper bound) \\
Paired-only                                & RNA           & WSI     & Same architecture restricted to the paired subset (Stages 1--2) \\
\bottomrule
\end{tabular}
\end{table}

% ── Bibliography ─────────────────────────────────────────────────────────────
\bibliography{bibliography_fixed}

\end{document}